%% file: main.tex
\documentclass[10pt,twocolumn,letterpaper]{article}
\usepackage[inline]{enumitem}
\usepackage{cvpr}              
\usepackage[dvipsnames]{xcolor}
\usepackage{tabularx}
\usepackage{tabu}
\usepackage{adjustbox}
\usepackage{comment,subcaption}
\usepackage{standalone}

\definecolor{cvprblue}{rgb}{0.21,0.49,0.74}
\usepackage[pagebackref,breaklinks,colorlinks,citecolor=cvprblue]{hyperref}
\usepackage[capitalize]{cleveref}
\usepackage{mathtools} 
\usepackage{amsmath} 
\usepackage[accsupp]{axessibility} 

\input{preamble}

\def\paperID{10226} 
\def\confName{CVPR}
\def\confYear{2024}

\title{ECoDepth: Effective Conditioning of Diffusion Models \\ for Monocular Depth Estimation}

\def\spaces{~~~~~~~~~~~~~~~~}

\author{
Suraj Patni\thanks{Equal contribution.}
\spaces{}~~
Aradhye Agarwal\footnotemark[1]
\spaces{} 
Chetan Arora\\
Indian Institute of Technology Delhi\\
\url{https://ecodepth-iitd.github.io}
}

\begin{document}
\maketitle


\input{sec/0_abstract}

\input{sec/1_intro}
\begin{figure*}[t]
	\centering
	\includegraphics[width=\textwidth, height=0.34\textwidth]{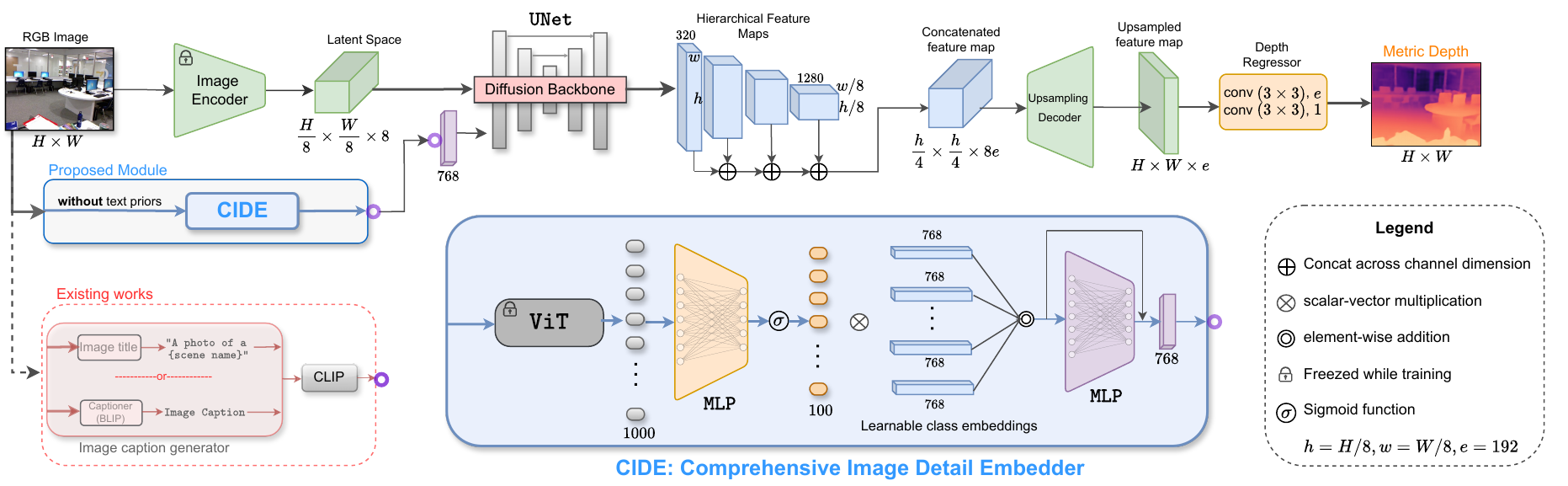} 
	\caption{\textbf{An overview of our proposed model:} The latent representation of the input image undergoes a diffusion process, which is conditioned by our proposed CIDE module. Within the CIDE module, the input image is fed through the frozen \vit model. From this, a linear combination of the learnt embeddings is computed, which is transformed to generate a 768-dimensional contextual embedding. This embedding is utilized to condition the diffusion backbone. Subsequently, hierarchical feature maps are extracted from the \unet's decoder which are concatenated and processed through a depth regressor to generate the depth map.}
	\label{fig:model_aarch}
\end{figure*}

\input{sec/2_related_work}
\input{sec/3_methodology}
\input{sec/4_5_experiments}

\input{sec/6_ablation}

\vspace{-2mm}
\input{sec/7_conclusion}
\input{sec/acknowledgement}

  
{
    \small
    \bibliographystyle{ieeenat_fullname}
    \bibliography{main}
}
\appendix
\input{sec/Supplementary}
\newpage
\end{document}

%% file: preamble.tex
\newcommand{\myfirstpara}[1]{\par \noindent \textbf{{#1}.}}
\newcommand{\mypara}[1]{\vspace{0.2em}\myfirstpara{#1}}

\def\side{\texttt{SIDE}\xspace}
\def\vit{\texttt{ViT}\xspace}
\def\vpd{\texttt{VPD}\xspace}
\def\tadp{\texttt{TADP}\xspace}
\def\zoe{\texttt{ZoEDepth}\xspace}
\def\sota{\texttt{SOTA}\xspace}
\def\clip{\texttt{CLIP}\xspace}
\def\nyu{\texttt{NYU Depth v2}\xspace}
\def\kitti{\texttt{KITTI}\xspace}
\def\newcrf{\texttt{NeWCRF}\xspace}
\def\sun{\texttt{Sun-RGBD}\xspace}
\def\ibims{\texttt{iBims1}\xspace}
\def\diode{\texttt{DIODE}\xspace}
\def\hypersim{\texttt{HyperSim}\xspace}

\def\mde{\texttt{MDE}\xspace}
\def\rde{\texttt{RDE}\xspace}
\def\blip{\texttt{BLIP-2}\xspace}
\def\laion{\texttt{LAION-400M}\xspace}
\def\lfms{\texttt{LFMs}\xspace}
\def\rmse{\texttt{RMSE}\xspace}
\def\cnns{\texttt{CNNs}\xspace}

\def\rgb{\texttt{RGB}\xspace}
\def\unet{\texttt{UNet}\xspace}
\def\mlp{\texttt{MLP}\xspace}
\def\cide{\texttt{CIDE}\xspace}
\def\ged{\texttt{GED}\xspace} 

\def\yb{\mathbf{y}}
\def\xb{\mathbf{x}}
\def\zb{\mathbf{z}}
\def\real{\mathbb{R}}
\def\P{\mathbb{P}}
\def\N{\mathcal{N}}
\def\Cb{\mathbf{\mathcal{C}}}
\def\U{\mathcal{U}}
\def\Eb{\mathbb{E}}
\def\Ec{\mathcal{E}}
\def\Ib{\boldsymbol{I}}
\def\eb{\boldsymbol{\epsilon}}

\definecolor{armygreen}{rgb}{0.29, 0.33, 0.13}
\definecolor{forestgreen}{rgb}{0.34, 0.90, 0.34}

%% file: sec/0_abstract.tex
\begin{abstract}

In the absence of parallax cues, a learning based single image depth estimation (\side) model relies heavily on shading and contextual cues in the image. While this simplicity is attractive, it is necessary to train such models on large and varied datasets, which are difficult to capture. It has been shown that using embeddings from pre-trained foundational models, such as \clip, improves zero shot transfer in several applications. Taking inspiration from this, in our paper we explore the use of global image priors generated from a pre-trained \vit model to provide more detailed contextual information. We argue that the embedding vector from a \vit model, pre-trained on a large dataset, captures greater relevant information for \side than the usual route of generating pseudo image captions, followed by \clip based text embeddings. Based on this idea, we propose a new \side model using a diffusion backbone which is conditioned on \vit embeddings. Our proposed design establishes a new state-of-the-art (\sota) for \side on \nyu dataset, achieving Abs Rel error of 0.059(14\% improvement) compared to 0.069 by the current \sota (\vpd). And on \kitti dataset, achieving Sq Rel error of 0.139 (2\% improvement) compared to 0.142 by the current \sota (\ged). For zero shot transfer with a model trained on \nyu, we report mean relative improvement of (20\%, 23\%, 81\%, 25\%) over \newcrf on (\sun, \ibims, \diode, \hypersim) datasets, compared to (16\%, 18\%, 45\%, 9\%) by \zoe. The code is available at \href{https://github.com/Aradhye2002/EcoDepth}{{this link}}.
\end{abstract}

%% file: sec/1_intro.tex
\begin{figure}
	\centering
	\begin{subfigure}{1.0\linewidth}
		\centering
		\includegraphics[width=1.0\linewidth]{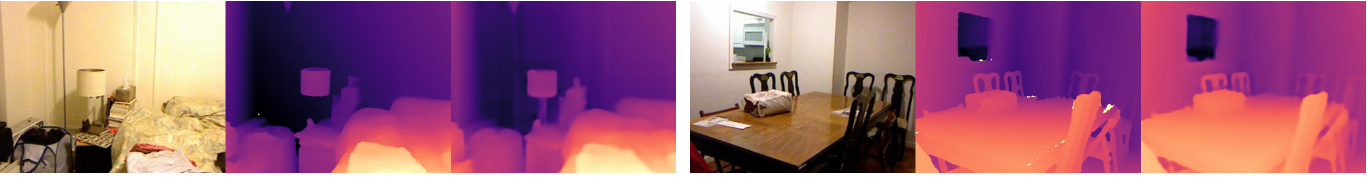}
		\caption{\sun\cite{sunrgbd_dataset}}
		\label{fig:sub1}
	\end{subfigure}\\
	\begin{subfigure}{1.0\linewidth}
		\centering
		\includegraphics[width=1.0\linewidth]{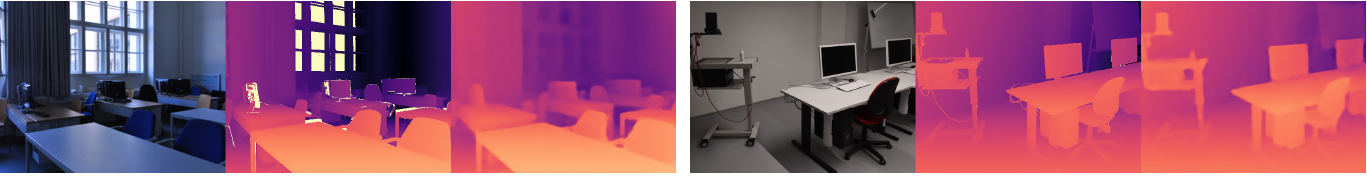}
		\caption{\ibims\cite{ibims_dataset}}
		\label{fig:sub2}
	\end{subfigure} \\
	\begin{subfigure}{1.0\linewidth}
	\centering
	\includegraphics[width=1.0\linewidth]{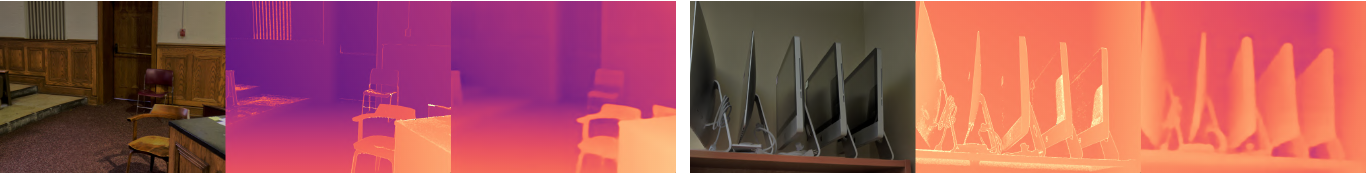}
	\caption{\diode\cite{diode_dataset}}
	\label{fig:sub1}
	\end{subfigure}\\
	\begin{subfigure}{1.0\linewidth}
		\centering
		\includegraphics[width=1.0\linewidth]{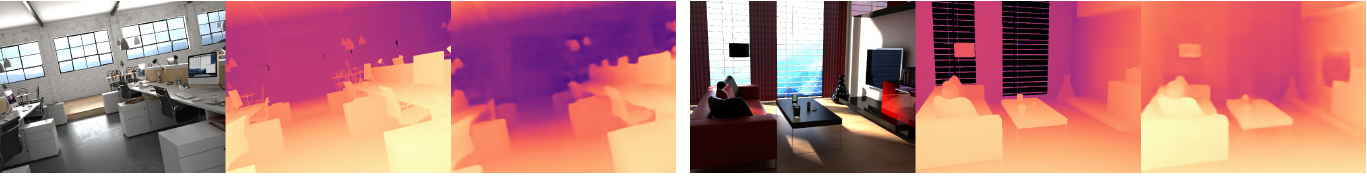}
		\caption{\hypersim\cite{hypersim_dataset}}
		\label{fig:sub2}
	\end{subfigure}
	\caption{Qualitative results across four different datasets, demonstrating the zero-shot performance of our model trained only on the \nyu dataset. Corresponding quantitative results are presented in Table \ref{tab:zero-shot-indoors}. The first column displays RGB images, the second column depicts ground truth depth, and the third column showcases our model's predicted depths. Additional images for each dataset are available in the Supplementary Material.}
	\label{fig:teasure}
\end{figure}

\section{Introduction}
\label{sec:intro}

Single Image Depth Estimation (\side) is the task of predicting per pixel depth using a single RGB image from a monocular camera. It is a fundamental problem in computer vision with applications in several domains, viz robotics, autonomous driving, and augmented reality. The problem is typically formulated in two flavors: metric depth estimation (\mde), and relative depth estimation (\rde). As the names suggest, \mde deals with estimation in physical units such as meters, whereas, \rde techniques focus on relative depth only, and require a per-image affine transformation as post-processing to convert to a physical unit.

Conventional geometric techniques for depth estimation typically rely on feature correspondence, parallax, and triangulation from two or more views. However, the problem becomes ill-posed for estimation from a single view. Intuitively, depth map is a 3D representation of the scene, whereas an RGB image is a 2D projection of the scene. Hence, it cannot be uniquely determined from a single RGB image. Therefore, learning-based \side models rely on visual cues like `shape from shading' and other contextual priors for per-pixel depth prediction. 

A data-driven approach makes the \side pipeline simpler, but also makes the learnt model dependent on the quality of training data. It has been observed that such models overfit on a particular training distribution/domain and fail to generalize on unseen data. This is especially true for \mde models when the range of depth in the training dataset is limited. Hence, training on multiple datasets, with wide variations in depth ranges has been proposed \cite{zoedepth,midas}.

On the other hand, development of large foundational models (\lfms) in recent years has altered the preferred design approach for many computer vision problems. These huge models are trained using extensive datasets of unlabeled images and learning objectives that are agnostic to specific tasks. The learnt embeddings from such pre-trained models have been shown to help generalization and zero-shot transfer in many applications. We are aware of at least two works for \mde in \side problem that have appeared in the last few months and make use of such foundational models. \vpd \cite{VPD} uses a text-to-image diffusion model pre-trained on \laion~\cite{laion400m_dataset} dataset having large-scale image-text pairs as the backbone. The model prompts the denoising \unet with textual inputs to make the visual contents interact with the text prompts. Since the problem formulation doesn't include text description as the input, the model generates simple descriptions such as \texttt{``A photo of a \{scene name\}''} based on the scene label given in the \nyu dataset. Another work, \tadp \cite{TADP} improves upon \vpd. Instead of simple image descriptions, \tadp uses \blip~\cite{BLIP} to generate image captions, and then uses \clip \cite{CLIP} embeddings of the pseudo-caption to condition the diffusion model. 

We view the above works as providing robust semantic context to \lfms for the actual task at hand, which helps in visual recognition in general, as well as \side. While we do agree with the broad motivations of these works, the question that we ask is if pseudo-captions are the most effective way to provide the semantic context. Textual descriptions of an image typically focus on large salient objects and emphasize on their relationships. On the other hand, large vision models for image classification, typically contain representation for even smaller objects present in the scene. Even when a single object is present in the scene, the representations typically capture uncertainties and ambiguities inherent in the scene. Hence, we posit that using embeddings from a transformer model, pretrained on a large image dataset unrelated to the \side task, captures more relevant information, and is a better alternative to using pseudo-captions' \clip embedding. Since diffusion-based models have shown their superiority for the dense prediction tasks in recent works \cite{DDP,depthgen,VPD,TADP,diffusiondepth}. Hence, we propose a diffusion backbone for our model, along with a novel \cide module which employs \vit \cite{vit_paper} to extract semantic context embeddings. These embeddings are subsequently utilized to condition the diffusion backbone.

\mypara{Contributions}
The key contributions of this work are: 
\begin{enumerate}[label=\textbf{(\arabic*)},leftmargin=*,topsep=0pt,itemsep=-1ex,partopsep=1ex,parsep=1ex]
\item We propose a new model for \mde in a \side task. The proposed model uses a conditional diffusion architecture with the semantic context being supplied through embeddings generated using a \vit model. 
Achieving a new state-of-the-art (\sota) performance, our method outperforms existing approaches on benchmark datasets, including the \nyu indoor and \kitti outdoor datasets. Notably, we report a significant improvement of 14\% in absolute relative error, achieving 0.059 compared to the current SOTA (\vpd) performance of 0.069 on \nyu. 
And we report an improvement of 2\% in square relative error, achieving 0.139 compared to the current SOTA (\ged) performance of 0.142 on \kitti. 
\item We show, qualitatively as well as quantitatively, that using \vit embeddings to provide semantic context is a better alternative to generating pseudo captions and then using its \clip embeddings to condition a \side model. In contrast to \tadp \cite{TADP}, which uses pseudo captions, but only achieves a \rmse of 0.225 on \nyu, we report a lower, and a new \sota, error of 0.218 (\vpd \cite{VPD} achieves 0.254).
\item We show that providing \vit conditioning, helps our model perform better in a zero shot transfer task. \zoe \cite{zoedepth}, the current \sota for zero-shot transfer, reports an improvement of (16\%, 18\%, 45\%, 9\%) over \newcrf on (\sun, \ibims, \diode, \hypersim) datasets, after training their model on 12 other datasets and \nyu. In contrast, we only train on \nyu, and report a much larger improvement of (21\%, 23\%, 81\%, 25\%).
\end{enumerate}

%% file: sec/2_related_work.tex
\section{Related Work}
\label{sec:related_work}

\myfirstpara{Traditional Methods}
Earlier techniques for \side have used Markov Random Fields \cite{saxena_etal}, non-parametric depth sampling \cite{karsch_etal}, and structural similarity with prior depth map \cite{herrera_etal} to predict pixel-wise depth.

\mypara{Deep Learning Techniques for \side}
Modern techniques have approached the problem as a dense regression problem. \cnns have been the dominant architecture for the \side in the last decade, with global-local network stack \cite{eigen2014depth}, and multi-scale \cite{laina2016deeper}, or encoder-decoder architecture \cite{fu2018deep} as some of the popular solution strategies. Recently, \texttt{PixelFormer} \cite{pixelformer} used a transformer-based encoder-decoder architecture with skip connections from encoders to decoders. \texttt{MIM} \cite{xie2023revealing} proposed masked image modeling as a general-purpose pre-training for geometric and motion tasks such as \side and pose estimation. Similarly, \texttt{AiT} \cite{ait} used mask augmentation and proposed soft tokens to generalize visual prediction tasks. While earlier works either focused on \mde for a specific dataset or \rde for generalization on multiple datasets, \zoe \cite{zoedepth} has proposed a generalized method for \mde that performs well in zero-shot transfer. We outperform \cite{zoedepth} by a large margin even when training on a single dataset - \nyu for indoor scenes or \kitti for outdoor scenes. Whereas, \cite{zoedepth} uses 12 datasets for pre-training and then fine-tunes on \nyu or \kitti for zero shot transfer.

\mypara{Diffusion-based Methods with Pretraining on Large Datasets}
Recently, many techniques for \side has used diffusion architectures. These techniques \cite{depthgen,DDP,VPD,TADP,diffusiondepth}, exploit prior knowledge acquired by pretraining on large datasets like \laion \cite{laion400m_dataset}, which consists of 400 million image-text pairs. In contrast, depth datasets, such as \nyu and \kitti, contain around 20-30 thousand image-depth pairs. \texttt{DepthGen} \cite{depthgen} and \texttt{DDP} \cite{DDP} work on a noise-to-depthmap paradigm and use images for conditional guidance of the diffusion process. \texttt{DepthGen} employs self-supervised pretraining on tasks like colorization, inpainting, and \texttt{JPEG} artifact removal, followed by supervised training on indoor and outdoor datasets \cite{scannet_dataset, scenenet_dataset, nyuv2, waymo_dataset, kitti}. \texttt{DDP} \cite{DDP} decouples the image encoder and map decoder, allowing the image encoder to run just once, while the lightweight map decoder is run multiple times. \vpd \cite{VPD} and \tadp \cite{TADP} use denoising \unet \cite{unet} as a backbone to extract the rich features at multiple scales. Also, they utilize text instead of image for conditioning the diffusion backbone.

\mypara{Vision Transformer for Scene Understanding}
The transformer architecture was initially proposed for \texttt{NLP} tasks \cite{vaswani2017attention}, but introduced to the computer vision community as Vision Transformer (\vit) in \cite{vit_paper}. Prior to this, \cnns dominated computer vision problems due to their ability to capture spatial hierarchies in image data. However, such architectures were constrained due to their ability to learn spatially localized features only. Transformer architecture has a weaker inductive bias and allows \vit to learn long-range dependencies, and robust, generalizable features. \vit architectures have replaced \cnns for \sota performance on most computer vision tasks in recent years. We use a pre-trained \vit model for providing semantic information to the diffusion backbone in our model. 

%% file: sec/3_methodology.tex
\section{Proposed Methodology}
\label{sec:methodology}

\subsection{Preliminaries}

\myfirstpara{Problem Formulation}
The objective of single image depth prediction task is to predict continuous values, denoted as $\yb \in \real^{H \times W}$, for every pixel present in the input \rgb image, $\xb \in [0,255]^{3 \times H \times W}$. Here $H$ and $W$ represent the height and width respectively of the input image.

\mypara{Diffusion Model}
Diffusion models are a class of generative models that progressively inject noise into the input data (forward pass) and then learn to reconstruct the original data in a reverse denoising process (reverse pass). There are three formulations of diffusion models: denoising diffusion probabilistic models (\texttt{DDPMs}) \cite{ho2020denoising}, score-based generative models \cite{song2019generative, song2020improved}, and those based on stochastic differential equations \cite{song2021maximum, song2020score}. \texttt{DDPMs} are of relevance to our paper, and are described below. The architecture of \texttt{DDPMs} consists of two Markov chains: a forward chain that adds noise to the data, and a reverse chain that converts noise back to data by learning transition kernels parameterized by deep neural networks. Formally, the forward pass is modeled as a Markov process:
\begin{equation}
\P \left( \zb_t \mid \zb_{t-1} \right) = \N \left( \zb_t; \sqrt{1-\beta_t} \zb_{t-1}, \beta_t \Ib \right).
\end{equation}
Here $\zb_t$ denotes the random variable at the $t^\text{th}$ time step, $\N(\zb; \boldsymbol{\mu}, \boldsymbol{\sigma})$ denotes Gaussian probability distribution, and $\beta_t$ is the noise schedule. The above equation leads to the analytic form of $\P(\zb_t \mid \zb_0), \forall t \in \{0, 1,\dots ,T\}$:
\begin{equation}
	\zb_t= \sqrt{\bar{\beta}_t} \zb_0 + \sqrt{1-\bar{\beta}_t} \eb,
	\label{eq:zt}
\end{equation}
where $\bar{\beta}_t = \prod_{s=1}^t \beta_s$ and $\eb \sim \N(\mathbf{0}, \Ib)$. The model then gradually removes noise by executing a learnable Markov chain in the reverse time direction, parameterized by a normal prior distribution $\P \left( \zb_{T} \right) = \N \left( \zb_{T} ; 0, \Ib \right)$ and a learnable transition kernel $\P_\theta \left( \zb_{t-1} \mid \zb_t \right)$ given by:
\begin{equation}
	\P_\theta \left( \zb_{t-1} \mid \zb_t\right) = \N \left( \zb_{t-1}; \mu_\theta \left( \zb_t, t \right), \Sigma_\theta \left( \zb_t, t \right) \right).
\end{equation}
The goal of the training process is to approximately match the reverse Markov chain with the actual time reversal of the forward Markov chain. Mathematically, parameter $\theta$ is adjusted so that the joint distribution of the reverse Markov chain $\P_\theta \left( \zb_0, \zb_1, \cdots, \zb_T \right) := \P \left( \zb_T \right) \prod_{t=1}^T \P_\theta \left( \zb_{t-1} \mid \zb_t \right)$ closely approximates that of the forward Markov chain $\P \left( \zb_0, \zb_1, \cdots, \zb_T\right) := \P \left( \zb_0 \right) \prod_{t=1}^T \P \left( \zb_t \mid \zb_{t-1} \right)$. This is achieved by minimizing the following loss:
\begin{equation}
\Eb_{t \sim \U[1,T], \zb_0 \sim \P \left( \zb_0 \right), \eb \sim \N(0,\Ib)}
\left[ 
	\left\| \eb - \eb_\theta \left( \zb_t, t \right) \right\|^2 
\right],
\end{equation}
where $\zb_t$ is computed using \cref{eq:zt}, and $\eb_\theta$ is predicted using a neural network, typically a \unet architecture \cite{unet}. In a conditional diffusion model $\eb_\theta \left( \zb_t, t \right)$ gets replaced by $\eb_\theta \left( \zb_t, t, \Cb \right)$, where $\Cb$ is a conditioning variable.

\subsection{Our Architecture}

\myfirstpara{Image Encoder and Latent Diffusion}
Diffusion models typically take large number of time steps to train, and are difficult to converge. Recently, \cite{latent-diffusion} proposed a new diffusion with improved convergence properties and is called ``stable-diffusion''. The key idea is to perform the diffusion in latent space, with latent embedding learnt separately through a variational autoencoder (VAE). The Encoder of VAE first transforms the input image $\xb$ of size $(H,W)$ to latent space, then we follow latent diffusion formulation and utilize the \unet used in Stable Diffusion\cite{latent-diffusion}. Utilizing latent diffusion formulation enables our architecture to capture multi-resolution features. Hence, we aggregate the feature maps from different layers of the \unet module (implementing conditional diffusion) by bringing them all to 1/4th resolution of the latent space, resulting in a feature map of size $8e \times H/32 \times W/32$.

\begin{figure}[t]
	\centering
	\begin{subfigure}{0.2\textwidth}
		\centering
		\includegraphics[height=\textwidth]{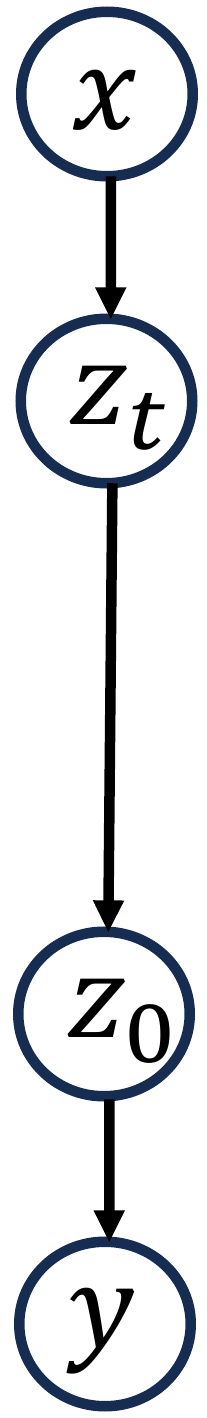}
		\caption{}
		\label{fig:graphical_model_a}
	\end{subfigure}
	\begin{subfigure}{0.2\textwidth}
		\centering
		\includegraphics[height=\textwidth]{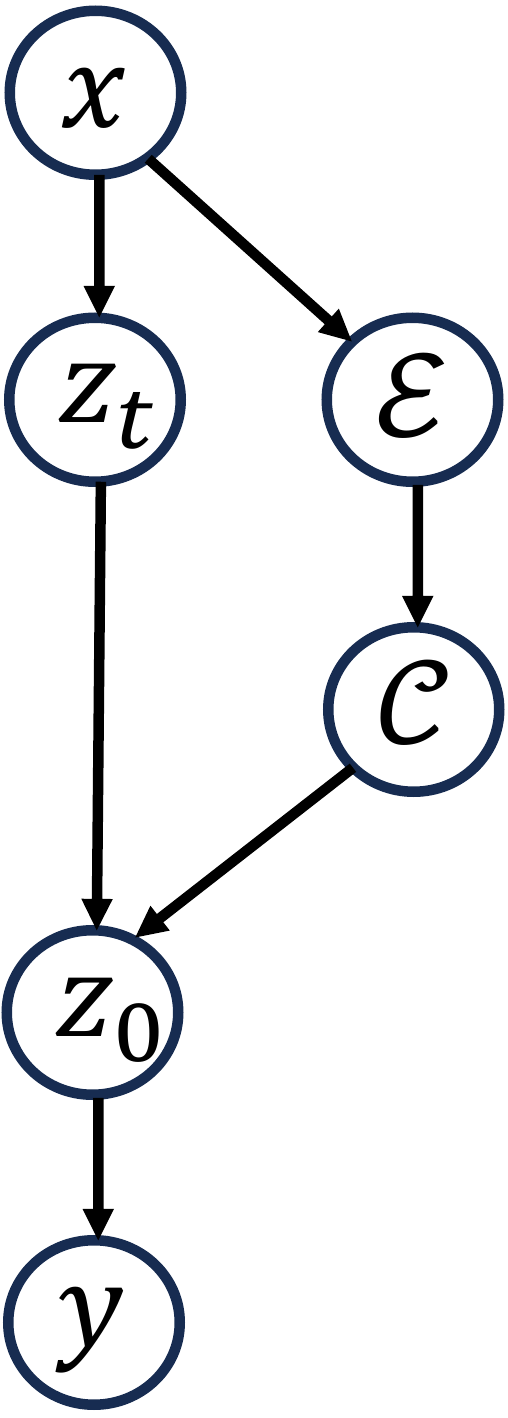}
		\caption{}
		\label{fig:graphical_model_b}
	\end{subfigure}	
 \caption{
 (a) Probabilistic graphical model corresponding to \vpd. (b) The same corresponding to our formulation. Here, $\Cb$ represents the semantic embedding derived from our \cide module. This embedding is internally generated by passing $\xb$ through the \vit, resulting in $\Ec$. Subsequently, $\Ec$ undergoes further processing to yield $\Cb$, which is then utilized in the conditional diffusion module implementing $\P(\zb_0 \mid \zb_t, \Cb)$. The output of the conditional diffusion module is fed into the Depth Regressor module within our architecture, implementing $\P(\yb \mid \zb_0)$.}
	\label{fig:graphical_model}
\end{figure}

\input{sec/nyu_table}
\input{sec/figure_qualitative_nyu}
\input{sec/kitti_table}
\input{sec/figure_qualitative_kitti}
\input{sec/table_zero_shot_transfer_indoor}
\input{sec/table_ablation_contextual_information}

\input{sec/figure_qualitative_ablation}

\mypara{Exploiting Semantic Context with Conditional Diffusion}
Recall that we formulated the single image depth estimation as a dense regression problem, predicting $\P(\yb \mid \xb)$, where $\yb$ and $\xb$ denote output depth, and input image respectively. In our diffusion formulation, we predict $\zb_0$ which is then used to predict pixel-wise depth, $\yb$. It has been shown that noise prediction in a diffusion model, $\eb_\theta$, can be seen as predicting the gradient of the density function, $\nabla_{\zb_t} \log \P(\zb_t)$. Hence, overall architecture for \side using a diffusion architecture can be seen as factorizing the conditional probability, as shown in the probabilistic graph model in \cref{fig:graphical_model_a}. To utilize additional semantic context generated from a \vit model, we condition it on the \vit embeddings as shown in \cref{fig:graphical_model_b}. Mathematically, we model:

\begin{align}
	\P ( \yb \mid \xb, \Ec) = & \P(\yb \mid \zb_0) \P(\zb_0 \mid \zb_t, \Cb) \P(\zb_t \mid \xb) \P(\Cb \mid \xb), \nonumber
	\shortintertext{where}
	\P(\Cb \mid \xb) = & \P(\Cb \mid \Ec) \P(\Ec \mid \xb).
	\label{eq:cide}
\end{align}

Here, the first term, $\P(\yb \mid \zb_0)$, is implemented through the Depth Regressor module explained earlier. Similarly, $\P(\zb_t \mid \xb)$ is implemented using the VAE's Encoder as described earlier. 
	We generate conditional information, $\Cb$ using our \emph{Comprehensive Image Detail Embedding} module (hereafter \cide, and explained below). The \cide module takes $\xb$ as input and generates embedding vector $\Cb$ (of dimension 768 in our design) as the output, thus, implementing $\P(\Cb \mid \xb)$ given in \cref{eq:cide}. 
	We use $\Ec$ to denote the embedding vector of a \vit module, and $\P(\Ec \mid \xb)$ is implemented through the \vit. $\P(\Cb \mid \Ec)$ is implemented using downstream modules in \cide consisting of learnable embeddings.
	The second term, $\P(\zb_0 \mid \zb_t, \Cb)$, is implemented using conditional diffusion,

\mypara{Comprehensive Image Detail Embedding (\cide) Module}
As described earlier, we believe using pseudo-captions to generate the semantic context has limited utility, as the textual descriptions typically focus on large salient objects only. Instead we propose our \cide module which use embeddings from a pre-trained \vit, and extract detailed semantic context from these embeddings. For this we take 1000 dimensional logit vector from \vit, and pass it through a two layer \mlp which converts it to a 100 dimensional vector. Subsequently, we employ this vector to compute the linear combination of 100 learnable embeddings. This resulting embedding undergoes a linear transformation to yield a semantic context vector of dimension 768, which is then passed to conditional diffusion module.

\mypara{Depth Regressor}
The output feature map undergoes through an Upsampling Decoder, comprised of deconvolution layers, followed by a Depth Regressor. The Depth Regressor is essentially a two-layer convolutional neural network (CNN), with the initial layer having dimensions Conv$(3\times3)$,192, and the subsequent layer Conv$(3\times3)$,1.

%% file: sec/nyu_table.tex
\begin{table*}[t]
\centering
\setlength{\tabcolsep}{10pt}
\caption{\textbf{Results on Indoor \nyu \cite{nyuv2} Dataset.} Results that are \textbf{bold} perform best. $\uparrow$ means the metric should be higher,  $\downarrow$ indicate lower is better. The evaluation uses an upper bound of 10 meters on the ground truth depth map. All the numbers for other works have been taken from the corresponding papers. For \texttt{MIM}, and \zoe we have used SwinV2-L 1K, and ZoeDepth-M12-N versions respectively. We see an overall improvement against \sota on all the metrics used for evaluation.}
\label{tab:nyu}
    \begin{tabular}{@{}llccccccc@{}}
        \toprule
        Method & Venue & Abs Rel$\downarrow$ & RMSE$\downarrow$ & ${\log_{10}\downarrow}$ & Sq Rel${\downarrow}$ & $\delta_1\uparrow$ & $\delta_2\uparrow$ & $\delta_3\uparrow$ \\
       \midrule
       Eigen et al.\cite{eigen2014depth} & NIPS’14 & 0.158 & 0.641 & - & -& 0.769 & 0.950 & 0.988 \\
       DORN\cite{DORN} & CVPR’18 & 0.115 & 0.509 & 0.051 & - & 0.828 & 0.965 & 0.992 \\
       SharpNet\cite{sharpnet} & ICCV'19 & 0.139 & 0.502 & 0.047 & - & 0.836 & 0.966 & 0.993 \\
       Chen et al.\cite{chen2019structure} & IJCAI-19 & 0.111 & 0.514 & 0.048 & - & 0.878 & 0.977 & 0.994 \\
       BTS\cite{bts_lee2019big} & Arxiv’19 & 0.110 & 0.392 & 0.047 & 0.066 & 0.885 & 0.978 & 0.994 \\
       AdaBins\cite{adabins} & CVPR’21 & 0.103 & 0.364 & 0.044 & - & 0.903 & 0.984 & 0.997 \\
       DPT\cite{DPT} & ICCV’21 & 0.110 & 0.357 & 0.045 & - & 0.904 & 0.988 & 0.998 \\
       P3Depth\cite{P3Depth} & CVPR’22 & 0.104 & 0.356 & 0.043 & - & 0.898 & 0.981 & 0.996 \\
       NeWCRFs\cite{newcrfs} & CVPR’22 & 0.095 & 0.334 & 0.041 & 0.045 & 0.922 & 0.992 & 0.998 \\
       SwinV2-B\cite{liu2022swin} & CVPR’22 & 0.133 & 0.462 & 0.059 & - & 0.819 & 0.975 & 0.995 \\
       SwinV2-L\cite{liu2022swin} & CVPR’22 & 0.112 & 0.381 & 0.051 & - & 0.886 & 0.984 & 0.997 \\
       Localbins\cite{bhat2022localbins} & ECCV'22 & 0.099 & 0.357 & 0.042 & - & 0.907 & 0.987 & 0.998 \\
       Jun et al.\cite{jun2022depth} & ECCV'22 & 0.098 & 0.355 & 0.042 & - & 0.913 & 0.987 & 0.998 \\
       PixelFormer\cite{pixelformer} & WACV'23 & 0.090 & 0.322 & 0.039 & 0.043 & 0.929 & 0.991 & 0.998 \\
       DDP\cite{DDP} & ICCV'23 & 0.094 & 0.329 & 0.040 & - & 0.921 & 0.990 & 0.998 \\
       MIM \cite{xie2023revealing} & CVPR'23 &  0.083 & 0.287 & 0.035 & - & 0.949 & 0.994 & 0.999 \\
       AiT\cite{ait} & ICCV'23 & 0.076 & 0.275 & 0.033 & - & 0.954 & 0.994 & 0.999 \\
       ZoeDepth \cite{zoedepth} & Arxiv’23 & 0.075 & 0.270 & 0.032 & 0.030 & 0.955 & 0.995 & 0.999 \\
       VPD\cite{VPD} & ICCV'23 & 0.069 & 0.254 & 0.030 & 0.027 & 0.964 & 0.995 & 0.999 \\
       \midrule
       Ours & CVPR'24 & \textbf{0.059} & \textbf{0.218} & \textbf{0.026} & \textbf{0.013} & \textbf{0.978} & \textbf{0.997} & \textbf{0.999} \\
       \bottomrule
    \end{tabular}
\end{table*}

%% file: sec/figure_qualitative_nyu.tex
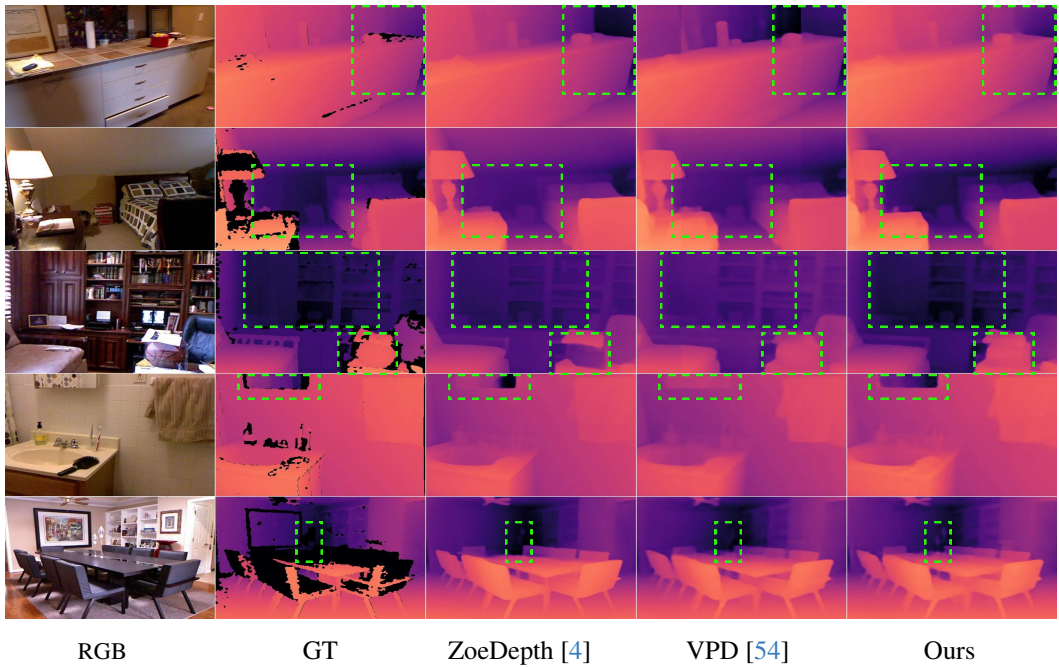
\begin{figure*}[!hbt]
	\centering
	\includegraphics[width=0.8\textwidth, height=0.362\textheight]{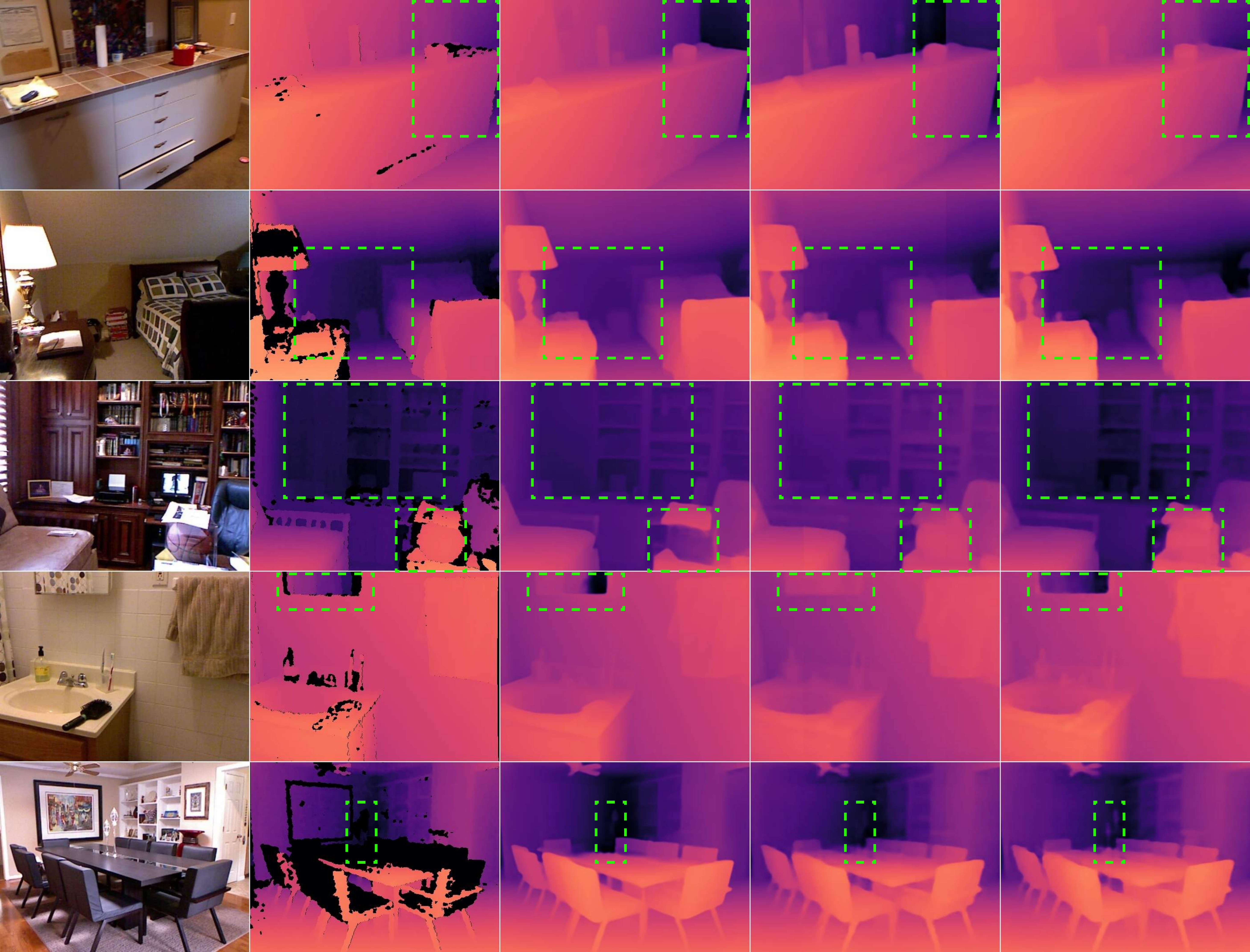}
	\put(-370, -15){\small RGB}
	\put(-285, -15){GT}
	\put(-230, -15){ZoeDepth \cite{zoedepth}}
	\put(-140, -15){VPD \cite{VPD}}
	\put(-50, -15){Ours}        
	\caption{\textbf{Visual Comparison on \nyu Indoor Dataset.} Note, our method's ability to delineate objects in terms of their depth, such as the table lamp in Row 5, even when such information is absent from the ground truth depth map.}
	\label{fig:qualtitative_nyu}
\end{figure*}

%% file: sec/kitti_table.tex
\begin{table*}[t] 
\centering
\setlength{\tabcolsep}{10pt} 
\caption{\textbf{Performance on the Outdoor \kitti \cite{kitti} Dataset.} Please refer to the caption of \cref{tab:nyu} for notation details. For \zoe results, we use the ZoeDepth-M12-K version following the authors' recommendation. In instances where results for certain methods were not reported in the respective works, denoted by ``-'', and the code is unavailable, we were unable to generate the missing numbers. Despite the saturation of results on the outdoor \kitti dataset, our method consistently achieves comparable or superior performance to the state-of-the-art (\sota) across all metrics. VPD\cite{VPD} cannot be trained on KITTI dataset as it required a per image text label which is not present in KITTI. The symbol $^\dagger$ indicates that the method utilizes additional information beyond RGB. }
\label{tab:kitti}
\begin{tabular}{@{}llccccccc@{}}
    \toprule
    Method & Venue & Abs Rel$\downarrow$ & Sq Rel$\downarrow$ &RMSE$_{log}\downarrow$ & RMSE$\downarrow$ & $\delta_1\uparrow$ & $\delta_2\uparrow$ & $\delta_3\uparrow$ \\
    \midrule
    Eigen et al.\cite{eigen2014depth} & NIPS’14 & 0.203 & 1.517 & 0.282 &  6.307 &  0.702 & 0.898 & 0.967 \\
    DORN\cite{DORN} & CVPR’18 & 0.072 & 0.307 & 0.120 & 2.727 & 0.932 & 0.984 & 0.994 \\
    BTS\cite{bts_lee2019big} & Arxiv’19 & 0.059 & 0.241 & 0.096 & 2.756 & 0.956 & 0.993 & 0.998 \\
    AdaBins\cite{adabins} & CVPR’21 & 0.067 & 0.190 & 0.088 & 2.960 & 0.949 & 0.992 & 0.998 \\
    DPT\cite{DPT} & ICCV’21 & 0.060 & - & 0.092 & 2.573 & 0.959 & 0.995 & 0.996 \\
    P3Depth\cite{P3Depth} & CVPR’22 & 0.071 & 0.270 & 0.103 & 2.842 & 0.953 & 0.993 & 0.998 \\
    NeWCRFs\cite{newcrfs} & CVPR’22 & 0.052 & 0.155 & 0.079 & 2.129 & 0.974 & 0.997 & 0.999 \\
    PixelFormer\cite{pixelformer} & WACV'23 & 0.051 & 0.149 & 0.077 & 2.081 & 0.976 & 0.997 & 0.999 \\
    ZoeDepth \cite{zoedepth} & Arxiv’23 & 0.054 & 0.189 & 0.083 & 2.440 & 0.97 & 0.996 & 0.999 \\
    DDP \cite{DDP} & ICCV'23 & 0.050 & 0.148 & 0.076 & 2.072 & 0.975 & 0.997 & 0.999 \\
    URCDC \cite{shao2023urcdc} & ToM'23 & 0.050 & 0.142 & 0.076 & 2.032 & 0.977 & 0.997 & 0.999 \\
    IEBins \cite{shao2023iebins} & NeurIPS'23 & 0.050 & 0.142 & 0.075 & 2.011 & 0.978 & 0.998 & 0.999 \\
    MIM \cite{xie2023revealing} & CVPR'23 & 0.050 & 0.139 & 0.075 & \textbf{1.966} & 0.977 & 0.998 & 1.000 \\
    GEDepth\cite{yang2023gedepth}$^\dagger$ & ICCV'23 & 0.048 & 0.142 & 0.076 & 2.044 & 0.976 & 0.997 & 0.999 \\
    \midrule
    Ours & CVPR'24 & \textbf{0.048} & \textbf{0.139} & \textbf{0.074} & 2.039 & \textbf{0.979} & \textbf{0.998} & \textbf{1.000} \\
    \bottomrule
\end{tabular}
\end{table*}

%% file: sec/figure_qualitative_kitti.tex
\begin{figure*}[t]
	\centering
	\includegraphics[width=1\textwidth, height=0.35\textwidth]{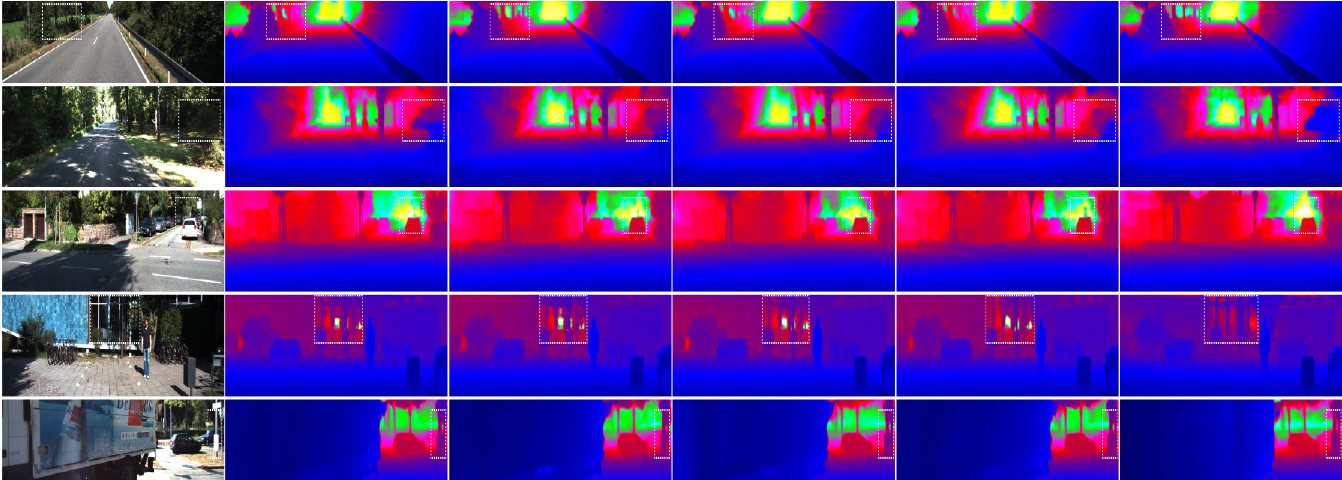}
	\put(-465, -15){\small RGB}
	\put(-400, -15) {NewCRF\cite{newcrfs}}
	\put(-315,-15) {Pixelformer\cite{pixelformer}}
	\put(-225,-15) {IEbins\cite{shao2023iebins}}
	\put(-160, -15) {URCDCdepth\cite{shao2023urcdc}}
	\put(-51,-15) {Ours}
	\caption{\textbf{Visual Comparison on \kitti Outdoor Dataset.} }
	\label{fig:qualtitative_kitti}
\end{figure*}
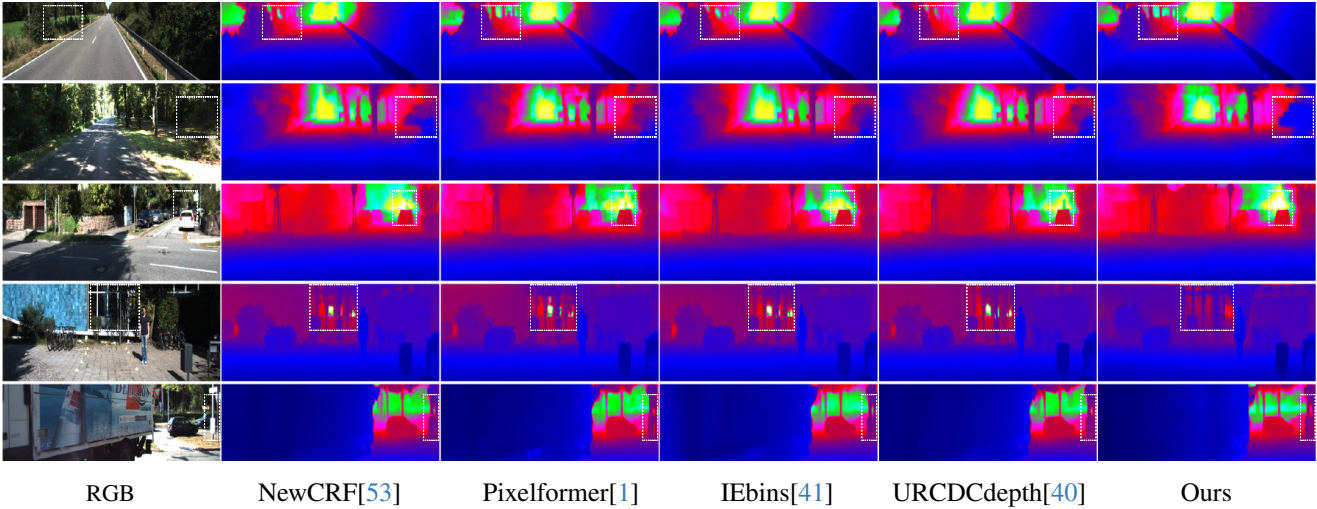
 

%% file: sec/table_zero_shot_transfer_indoor.tex
\begin{table*}[t]
\footnotesize
\centering
\setlength{\tabcolsep}{3pt} 
\caption{\textbf{Quantitative results for zero-shot transfer to four unseen indoor datasets.} mRI$_\theta$ denotes the mean relative improvement with respect to NeWCRFs across all metrics ($\delta_1$, REL, RMSE). Evaluation depth is capped at 8m for SUN RGB-D, 10m for iBims and DIODE Indoor, and 80m for HyperSim. Best results are in bold, second best are underlined. Our mRI$_\theta$  outperforms all methods across all datasets by a large margin. $^\dagger$ denotes that ZoeD is trained on 12 datasets and our method is trained only on NYUv2.}
\label{tab:zero-shot-indoors}
\begin{tabular}
{@{}
    l@{\hspace{6pt}}
    c@{\hspace{6pt}}c@{\hspace{3pt}}c@{\hspace{3pt}}|c@{\hspace{3pt}}|
    c@{\hspace{6pt}}c@{\hspace{3pt}}c@{\hspace{3pt}}|c@{\hspace{3pt}}|
    c@{\hspace{6pt}}c@{\hspace{3pt}}c@{\hspace{3pt}}|c@{\hspace{3pt}}|
    c@{\hspace{6pt}}c@{\hspace{3pt}}c@{\hspace{3pt}}|c@{\hspace{3pt}}
    @{}}
\toprule
& \multicolumn{4}{c|}{\textbf{SUN RGB-D}} & \multicolumn{4}{c|}{\textbf{iBims-1 Benchmark}} & \multicolumn{4}{c|}{\textbf{DIODE Indoor}} & \multicolumn{4}{c}{\textbf{HyperSim}}
\\
Method & $\delta_{1}$\,$\uparrow$ & REL\,$\downarrow$ & RMSE\,$\downarrow$ & mRI$_\theta$\,$\uparrow$ & $\delta_{1}$\,$\uparrow$ & REL\,$\downarrow$ & RMSE\,$\downarrow$ & mRI$_\theta$\,$\uparrow$ & $\delta_{1}$\,$\uparrow$ & REL\,$\downarrow$ & RMSE\,$\downarrow$ & mRI$_\theta$\,$\uparrow$ & $\delta_{1}$\,$\uparrow$ & REL\,$\downarrow$ & RMSE\,$\downarrow$ & mRI$_\theta$\,$\uparrow$   
\\ 
\midrule
BTS~\cite{bts_lee2019big} & 0.740 & 0.172& 0.515 & -14.2\% & 0.538 & 0.231& 0.919 & -6.9\%  & 0.210 & 0.418& 1.905 & 2.3\%& 0.225 & 0.476& 6.404 & -8.6\% 
\\
AdaBins~\cite{adabins}  & 0.771 & 0.159& 0.476 & -7.0\%  & 0.555 & 0.212& 0.901 & -2.1\%  & 0.174 & 0.443& 1.963 & -7.2\%  & 0.221 & 0.483& 6.546 & -10.5\%
\\
LocalBins~\cite{bhat2022localbins} & 0.777 & 0.156& 0.470 & -5.6\%  & 0.558  & 0.211 & 0.880  & -0.7\%  & 0.229  & 0.412 & 1.853 & 7.1\% & 0.234  & 0.468 & 6.362 & -6.6\%  
\\
NeWCRFs~\cite{newcrfs} & 0.798 & 0.151& 0.424 & 0.0\%& 0.548 & 0.206& 0.861 & 0.0\% & 0.187 & 0.404& 1.867 & 0.0\%& 0.255 & \underline{0.442}& 6.017 & 0.0\%  
\\
VPD \cite{VPD} & 0.861 & 0.121 & 0.355 & 14.7\% & 0.627 & 0.187 & 0.767 & 11.5\% & \underline{0.480} & 0.392 & \underline{1.295} & \underline{63.4\%} & \underline{0.333} & 0.531 & 5.111 & 8.5\% 
\\
ZoeD-M12-N$^\dagger$ \cite{zoedepth}  & \underline{0.864} & \underline{0.119} & \underline{0.346} &     \underline{16.0}\% & \underline{0.658} &   \underline{0.169} &    \underline{0.711} & \underline{18.5\%} & 0.376 &  \textbf{0.327} &   1.588 &  45.0\% & 0.292 &  \textbf{0.410} &   5.771 &      \underline{8.6\%} 
\\

\midrule
Ours & \textbf{0.885} & \textbf{0.112} & \textbf{0.319} & \textbf{20.5\%} & \textbf{0.688} & \textbf{0.163} & \textbf{0.664} & \textbf{23.1\%} & \textbf{0.545} & \underline{0.344} & \textbf{1.164} & \textbf{81.3\%} & \textbf{0.394} & \underline{0.442} & \textbf{4.739} & \textbf{25.2\%} 
\\
\bottomrule
\end{tabular}
\end{table*}

%% file: sec/table_ablation_contextual_information.tex
\begin{table}[t]
	\centering
	\caption{Effectiveness of different embeddings for guiding the diffusion process for depth estimation on \nyu dataset. In the third row, rather than using pseudo caption embeddings generated from the scene label (as implemented by \vpd \cite{VPD}), we provide the one-hot vector representing the scene label as a condition to the diffusion model. We observe a slight improvement in the metrics, highlighting that information content in the caption is similar to that in one-hot label.}
	\label{tab:ablation_embeddings_benefit}
	\begin{tabular}{lccc}
		\toprule
		Embeddings & RMSE$\downarrow$ & Abs Rel$\downarrow$ & $\delta_1\uparrow$  \\
		\midrule
		Scene label emb. \cite{VPD} & 0.254 & 0.069 & 0.964 \\
		Text caption emb. \cite{TADP} & 0.225 & 0.062 & 0.976 \\
		One-hot vector emb. & 0.244 & 0.067 & 0.968 \\
		Proposed scene emb. & \textbf{0.218} & \textbf{0.059} & \textbf{0.978} \\
		\bottomrule
	\end{tabular}
\end{table}

%% file: sec/figure_qualitative_ablation.tex
\begin{figure*}[t]
	\centering
	\includegraphics[width=0.9\linewidth]{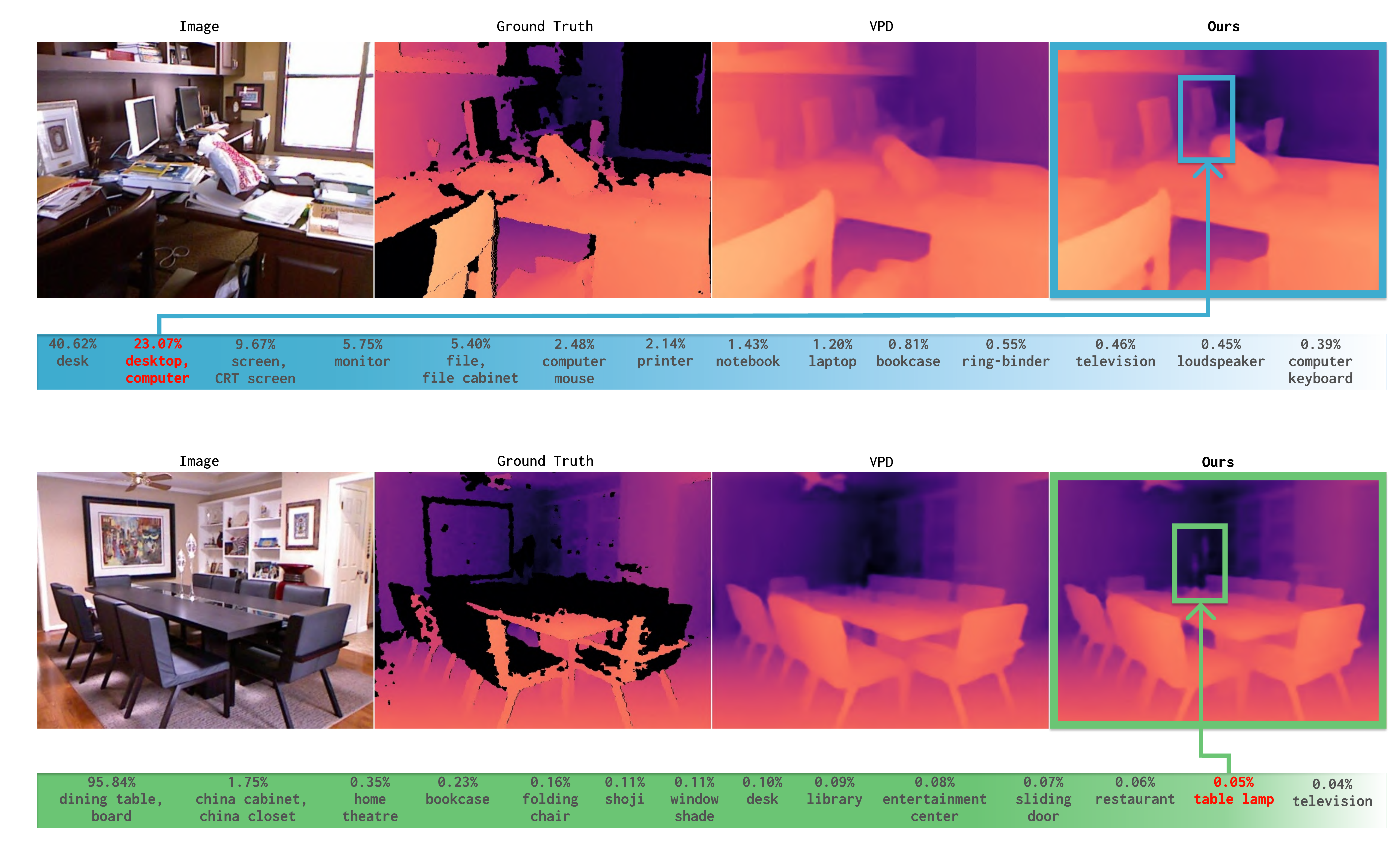}
	\captionof{figure}{ Visualization of improvements over \vpd \cite{VPD} in our model due to \vit embeddings passed down as conditional vectors (in \texttt{\textcolor{blue}{blue}} and \texttt{\textcolor{green}{green}}). In the above images, \vit detects the desktop computer (first image) and table lamp (second image) with high probability, and they are thus better detected by our model. Additional visualizations are provided in the supplementary material.}
	\label{fig:vit_logits_viz}
\end{figure*}

%% file: sec/4_5_experiments.tex
\section{Experiments and Results}
\label{sec:exp}

\myfirstpara{Datasets and Evaluation}
We use \nyu \cite{nyuv2} and \kitti \cite{kitti} as the primary datasets for training. The \nyu dataset is a widely used indoor benchmark for monocular depth estimation, containing over 24k densely labeled pairs of \rgb and depth images in the train set and 654 in the test set. The dataset covers a wide range of indoor scenes and includes challenging scenarios such as reflective surfaces, transparent objects, and occlusions. The ground truth depth maps are obtained using a structured light sensor and are provided at a resolution of $640 \times 480$. \kitti dataset is a widely used outdoor benchmark for monocular depth estimation, containing over 24k densely labeled pairs of \rgb and depth images. The dataset covers outdoor driving scenarios and includes varying lighting conditions, weather, and occlusions. The ground truth depth maps are obtained using a \texttt{Velodyne} {LiDAR} sensor and are provided at a resolution of $1242 \times 375$. To demonstrate generalizability, we evaluate zero-shot performance on the following datasets: \sun \cite{sunrgbd_dataset}, \ibims \cite{ibims_dataset}, \diode \cite{diode_dataset}, and \hypersim \cite{hypersim_dataset}. Whereas, the main paper contains mostly quantitative, and only some representative visual results, detailed visual results on each dataset are included in the supplementary.

\mypara{Implementation Details}
Our model is implemented using PyTorch \cite{pytorch}. For optimization, we have used  AdamW optimizer \cite{adamw} with $\beta_0$ values of 0.9 and 0.999, a batch size of 32, and a weight decay of $0.1$. We train our model for 25 epochs for both \kitti and \nyu datasets, with an initial learning rate of $3 \times 10^{-5}$. 
We first linearly increase learning rate to $5 \times 10^{-4}$, and then linearly decrease across training iterations. We use usual data augmentation techniques, including random hue addition, horizontal flipping, changing the image brightness, and Cut Depth \cite{cutdepth}. Our model takes approximately 21 minutes per epoch to train using 8 NVIDIA A100 GPUs.

\subsection{Comparison on Benchmark Datasets}

\myfirstpara{Comparison on \nyu}
\cref{tab:nyu} shows the comparison of our proposed method with \sota methods on the indoor \nyu dataset \cite{nyuv2}. We achieve a new state of the art on this dataset. Our methods perform better than the previous \sota (\cite{VPD})) by a large margin of 14\% in terms of \rmse. \cref{fig:qualtitative_nyu} provides a qualitative comparison on the dataset.

\mypara{Comparison on \kitti}
\cref{tab:kitti} shows the comparison with various methods on the \kitti dataset \cite{kitti}. \cref{fig:qualtitative_kitti} shows the qualitative results. Unlike \nyu dataset, \kitti is an outdoor dataset. On this dataset also, we achieve similar or better performance than all existing state-of-the-art techniques. 

\subsection{Generalization and Zero Shot Transfer}
Unlike state of the art on zero short transfer (\zoe \cite{zoedepth}), which requiring training on many datasets (12 in their case) for effective zero shot transfer, we show that our model generalizes well to other unseen dataset even when trained on a single \nyu dataset. \cref{tab:zero-shot-indoors} shows quantitative results to back our findings.

%% file: sec/6_ablation.tex
\subsection{Ablation Study}

\myfirstpara{Effect of Contextual Information}
As highlighted in the motivation, a key observation of this study is the richness of information contained within the output probability vector of the \vit, surpassing the textual embeddings employed in current state-of-the-art (e.g., \vpd\cite{VPD}). To compare with the utilization of scene label information (as implemented by \vpd), we construct the conditioning embedding as a one-hot vector using the scene label, and subsequently transform it using an MLP. As illustrated in \cref{tab:ablation_embeddings_benefit}, we observe a slight improvement over \vpd, indicating that 
the information content in pseudo-captions resembles that of one-hot labels. Furthermore, our proposed method surpasses both the approaches.

\mypara{Qualitative Results}
Perhaps the most important task would be to verify that use of rich probability vector instead of text embeddings actually results in an improvement in depth as a direct consequence. We do this by considering the top few objects predicted by the \vit and correlating this with the depth predicted at these objects. We hypothesise when a particular class say \emph{dog} is predicted with a high probability (hence there must be a dog in the image), then the corresponding depth must also be more accurate. We show this in \cref{fig:vit_logits_viz}.

%% file: sec/7_conclusion.tex
\section{Conclusion}
\label{sec:conclusion}

We presented a new architecture block Comprehensive Image Detail Embedding (\cide)  module for robust monocular depth estimation in this paper. Our key idea is to highlight the limitations of using pseudo-captions to provide contextual information, and instead propose to use richer class-wise probability generated by a classification model, such as \vit. The motivation is that, while textual embedding typically highlight salient objects, class-wise probability vector preserves more details, including smaller objects in the background also. We implement the idea using proposed \cide module cascaded with a conditional diffusion pipeline for monocular depth estimation. We demonstrated the effectiveness of our approach on several benchmark datasets and showed that it outperforms \sota methods by a significant margin. 

%% file: sec/acknowledgement.tex
\mypara{Acknowledgement}
We thank Kartik Anand for assistance with some of the experiments. This work has been supported by funding through Staqu Technologies Pvt. Ltd., and Department of Science and Technology (DST), Government of India. The travel for Suraj Patni is supported through Research Acceleration funds of Department of Computer Science and Engineering, IIT Delhi. Aradhye Agarwal's travel is made possible by courtesy of Google.

%% file: sec/Supplementary.tex
\maketitlesupplementary

\section{Ablation Study}
\subsection{Effect of \textbf{\vit} Architecture}
Table \ref{tab:effect_of_ViT} investigates the impact of varying \vit sizes on the generation of embeddings from RGB images. Our results for the \nyu \cite{nyuv2} dataset suggest that ViT-base yields optimal performance. Additionally, our observations in the \kitti dataset align with a similar trend.

\begin{table}[h]
  \centering
  \caption{\textbf{Ablation Study on ViT Sizes:} Performance comparison of different ViT variants in terms of parameters and depth error metrics on the NYUv2 \cite{nyuv2} dataset. The results guide the selection of ViT-base in our final architecture. Best results are in \textbf{bold}.}

  \begin{tabular}{@{}lcccc@{}}
    \toprule
    Classifier & $\#$Parameters & RMSE$\downarrow$ & Abs Rel$\downarrow$ & $\delta_1\uparrow$\\
    \midrule
    ViT-base & 86.6 M & \textbf{0.218} & \textbf{0.059} & \textbf{0.978} \\
    deit-base & 86.6 M & 0.218 & 0.059 & 0.978 \\
    ViT-large & 303.3 M & 0.218 & 0.060 & 0.978 \\
    ViT-huge & 630.8 M & 0.219 & 0.060 & 0.978 \\
    
    \bottomrule
  \end{tabular}
  \label{tab:effect_of_ViT}
\end{table}

\begin{table}[h]
\centering
\caption{\textbf{Ablation Study on dimension of Learnable Scene Embeddings (N):} The table shows the impact of varying the dimension of learnable scene embeddings on the depth error metrics. We observe a decrease in error with increasing N until saturation occurs at N=100, prompting us to limit the model parameters to N=100. Best results are highlighted in \textbf{bold}.}

\begin{tabular}{@{}lcccc@{}}
    \toprule
    N & RMSE$\downarrow$ & Abs Rel$\downarrow$ & log$_{10}\downarrow$ & $\delta_1\uparrow$\\
    \midrule
    10 & 0.219 & 0.061 & 0.027 & 0.978 \\
    50 & 0.219 & 0.060 & 0.026 & 0.978 \\
    100 & \textbf{0.218} & \textbf{0.059} & \textbf{0.026} & \textbf{0.978} \\
    200 & 0.218 & 0.060 & 0.026 & 0.978 \\
    \bottomrule
\end{tabular}
\label{tab:effect_of_N}
\end{table}

\subsection{Additional Qualitative Ablation}
\label{supp:vit_logits_viz}
In Fig. \ref{fig:sup_vit_logits_viz}, we present supplementary qualitative ablation results that highlight the correlation between value of ViT logits and the improvement in the predicted depth. The visualization demonstrates that elevated value of ViT logits for specific objects contribute to our model's ability to focus on those objects, enhancing the accuracy of predicted depth in corresponding regions.

\section{Architectural Details}
\subsection{Image Encoder}
Similar to Latent Diffusion \cite{latent-diffusion}, we employed the VQVAE's encoder to transition from image space to latent space.
\subsection{Upsampling Decoder}
\begin{figure}[h]
	\centering
	\includegraphics[width=0.4\textwidth]{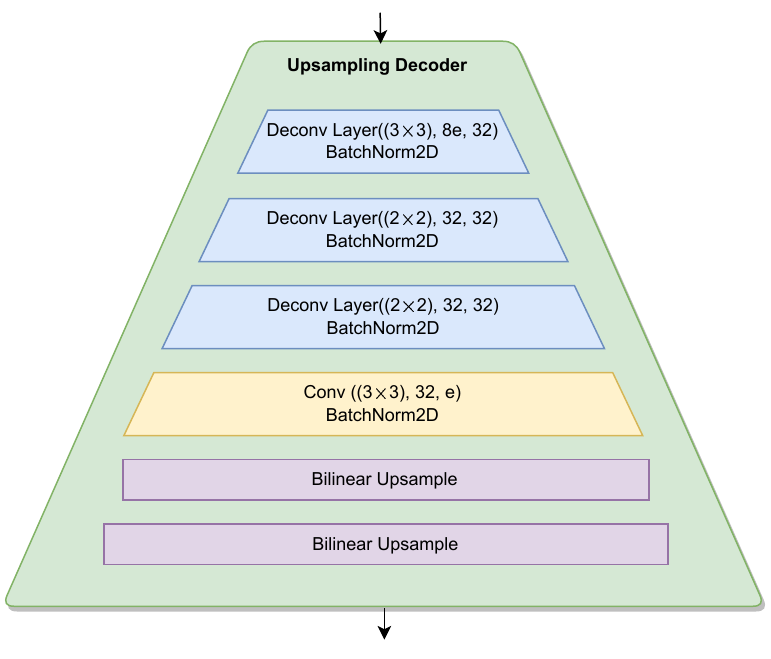}
        \caption{Detailed architecture of the upsampling decoder, responsible for upsampling the concatenated feature map to obtain the final feature map at a resolution of $H \times W$, $e=192$}
	\label{fig:decoder_aarch}
\end{figure}

After obtaining the hierarchical feature map from denoising UNet, the concatenated feature map undergoes upsampling, transitioning from a resolution of $64 \times 64$ back to $H \times W$. Refer to Fig. \ref{fig:decoder_aarch} for a detailed view of the upsampling decoder architecture.

\input{sec/sup_figure_qualitative_ablation}
\input{sec/sup_hyperparameter_table}

\section{Additional Experimental Details}
\subsection{Hyperparameters}
For reproducibility of the results presented in the main paper and the supplementary material, we provide a comprehensive list of the hyper parameters employed in our experiments in Table \ref{tab:hyperparameters}.


\input{sec/sup_all_datasets_qualit_comp.tex}
\section{Qualitative Results for Zero-Shot Performance Across Datasets}
\label{supp:qualit_all_datasets}
In the main paper, we presented a quantitative comparison of our method's zero-shot performance. Here, we provide a qualitative assessment of our method's performance in comparison to \zoe \cite{zoedepth} across the \hypersim, \diode, \sun and \ibims datasets in Fig. \ref{fig:qualtitative_hypersim}, \ref{fig:qualtitative_diode}, \ref{fig:qualtitative_sun} and \ref{fig:qualtitative_ibims}.

\clearpage
\clearpage
    

%% file: sec/sup_figure_qualitative_ablation.tex
\begin{figure*}[h]
	\centering
	\includegraphics[width=0.9\linewidth,height=0.6\linewidth]{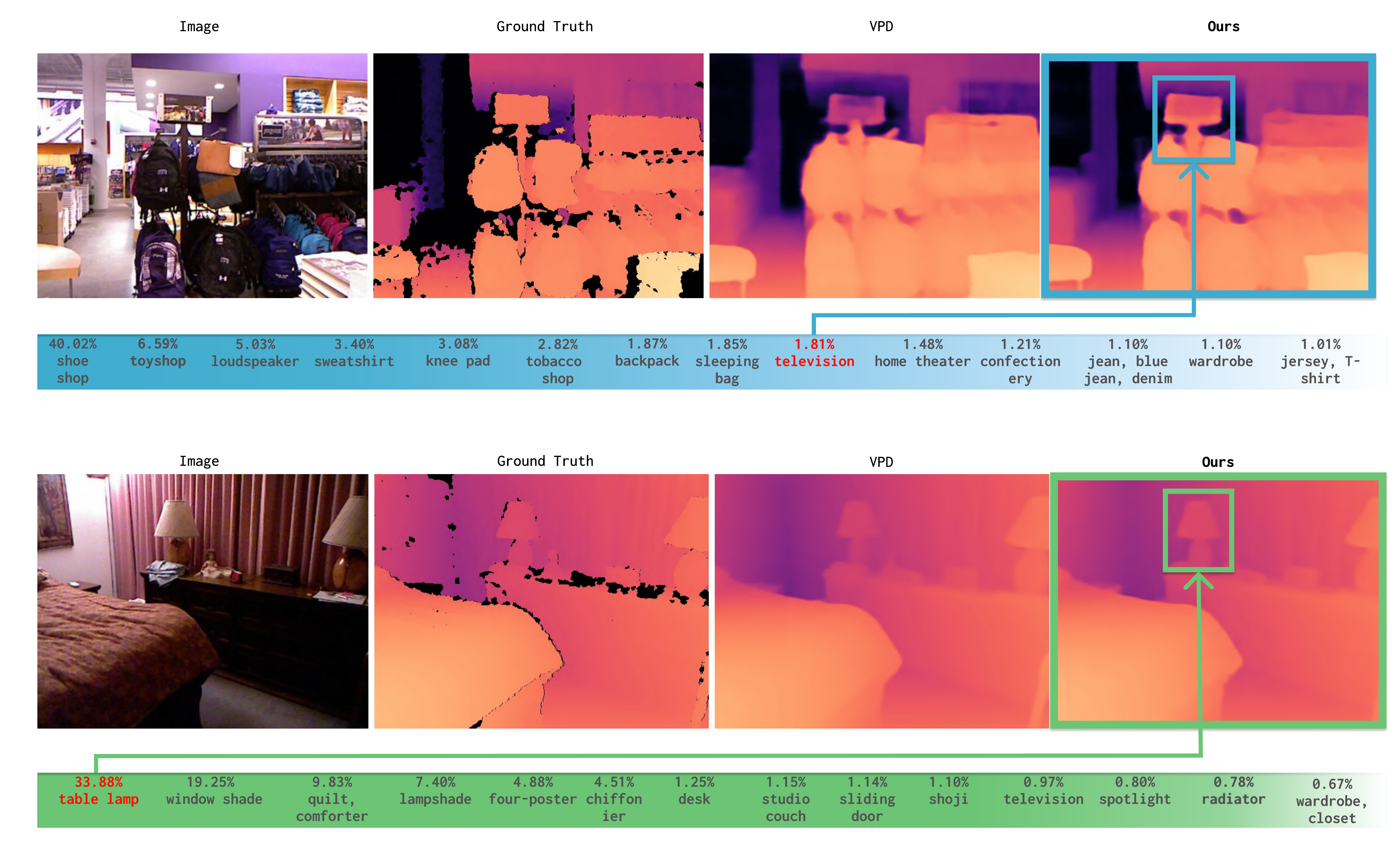}
    \captionof{figure}{Enhanced visualizations showcasing improvements over \vpd \cite{VPD} in our model, facilitated by \vit embeddings employed as conditional vectors for the denoising procedure. In the presented images, our model demonstrates heightened accuracy in detecting objects, such as the television (\texttt{\textcolor{blue}{blue}} in first image) and table lamp (\texttt{\textcolor{green}{green}} in second image) when these are detected with high probability by ViT.}
    \vspace{-10pt}
	\label{fig:sup_vit_logits_viz}
\end{figure*}

%% file: sec/sup_hyperparameter_table.tex
\begin{table}[!htpb]
    \centering
    \caption{Hyper-parameter settings for our model.}
    \begin{tabular}{@{}lr@{}}
        \toprule
        Hyper-parameter & Value \\
        \midrule
        Learning rate schedule & one cycle \\
        Min learning Rate & $3 \times 10^{-5}$\\
        Max learning Rage & $5 \times 10^{-4}$ \\
        Batch Size & $32$\\
        Optimizer & AdamW~\cite{adamw}\\
        $\beta_s$ in optimizer & ($0.9$, $0.999$) \\
        Weight Decay & $0.1$\\
        Layer Decay Rate & $0.9$ \\
        Embedding Dimension & $192$ \\
        Variance focus in SiLog loss & $0.85$ \\
        ViT Size  &  ViT-base \\
        Number of learnable emb. & 100 \\
        epochs & 25 \\
        \bottomrule
    \end{tabular}
    \label{tab:hyperparameters}
\end{table}

%% file: sec/sup_all_datasets_qualit_comp.tex
\begin{figure*}[!h]
\centering
\hspace{1cm}
\begin{picture}(0,0)
    \put(-34,135){\rotatebox{0}{RGB}}
    \put(-34,95){\rotatebox{0}{GT}}
    \put(-34,55){\rotatebox{0}{Zoed}}
    \put(-34,15){\rotatebox{0}{Ours}}
\end{picture}
\includegraphics[width=0.93\textwidth]{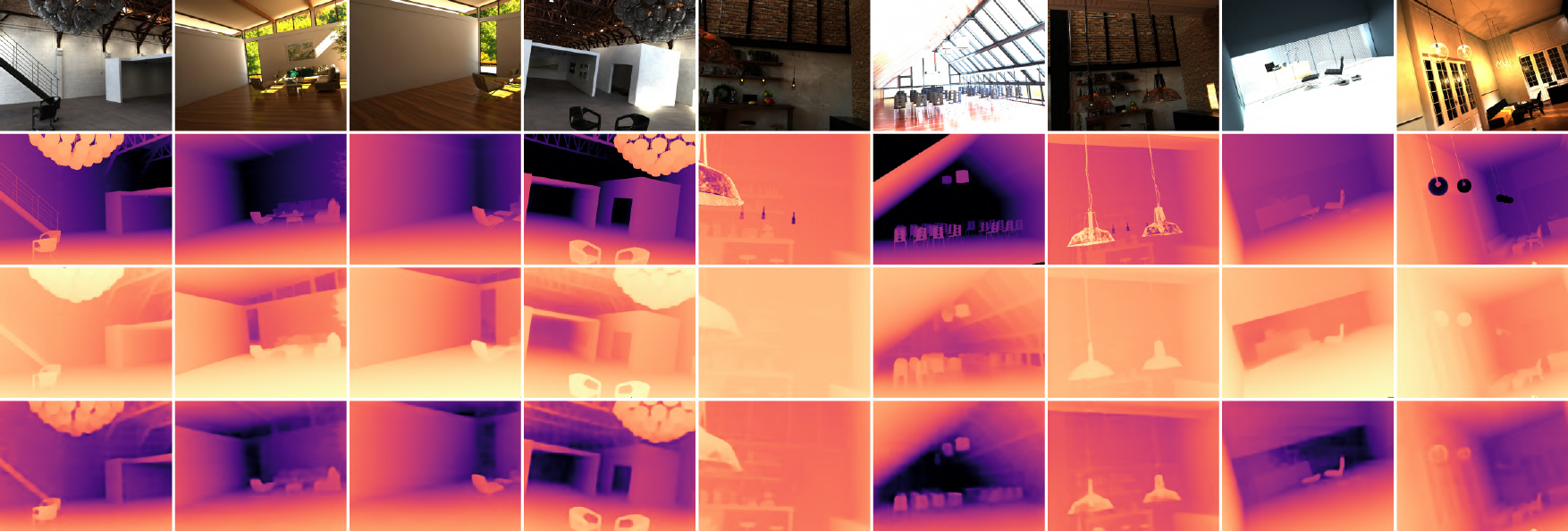}\vspace{1cm}

\hspace{1cm}
\begin{picture}(0,0)
    \put(-34,135){\rotatebox{0}{RGB}}
    \put(-34,95){\rotatebox{0}{GT}}
    \put(-34,55){\rotatebox{0}{Zoed}}
    \put(-34,15){\rotatebox{0}{Ours}}
\end{picture}
\includegraphics[width=0.93\textwidth]{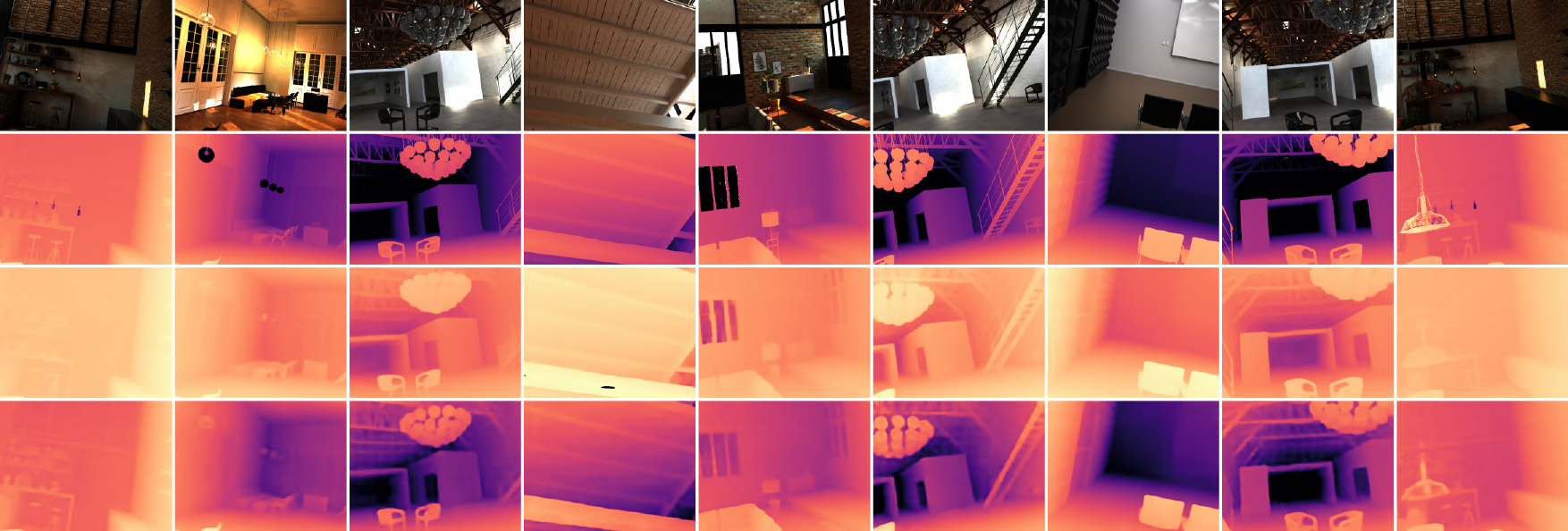}
\caption{\textbf{Qualitative Comparison on the \hypersim~\cite{hypersim_dataset} Dataset.} 
Our depth predictions are contrasted with those of Zoedepth\cite{zoedepth}. 
The first row displays RGB images, the second row shows groundtruth depth, 
the third row exhibits Zoedepth\cite{zoedepth}'s depth, and the fourth row showcases our depth predictions. 
To facilitate visual comparison, the colormap scale remains consistent across corresponding depth maps. 
Our model, trained only on \nyu, is compared with Zoedepth\cite{zoedepth}, which is trained on 12 datasets and then fine-tuned on \nyu.}
\label{fig:qualtitative_hypersim}
\end{figure*}

\begin{figure*}[h]
\centering
\hspace{1cm}
\begin{picture}(0,0)
    \put(-34,135){\rotatebox{0}{RGB}}
    \put(-34,95){\rotatebox{0}{GT}}
    \put(-34,55){\rotatebox{0}{Zoed}}
    \put(-34,15){\rotatebox{0}{Ours}}
\end{picture}
\includegraphics[width=0.93\textwidth]{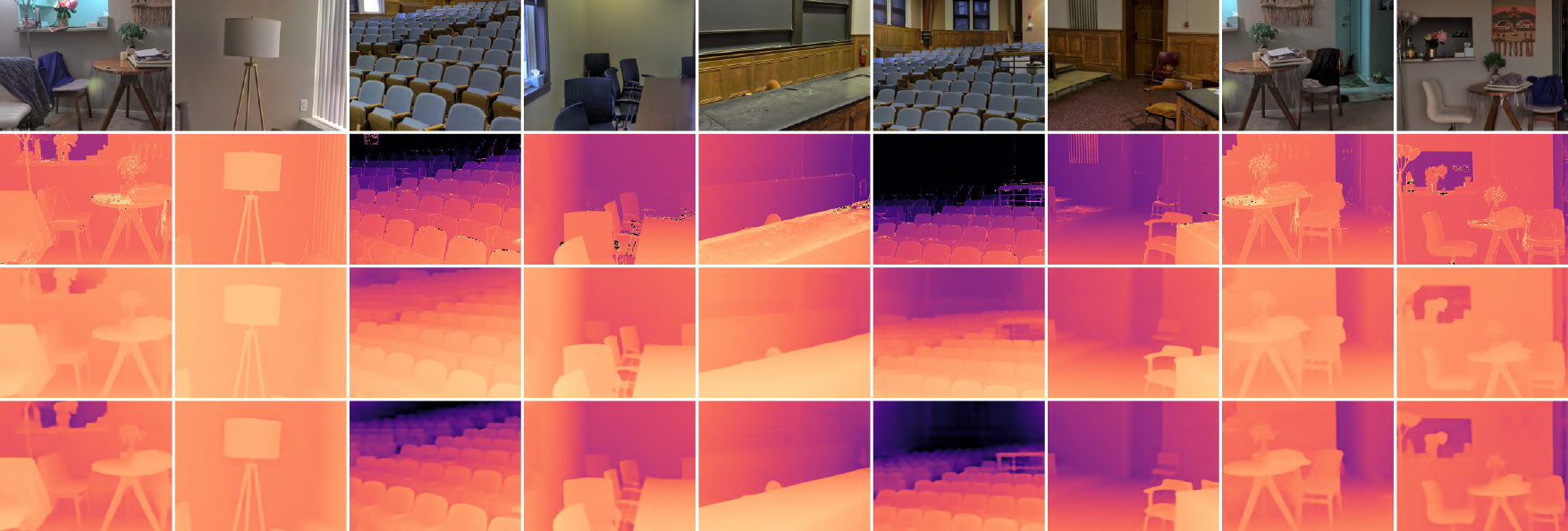}\vspace{1cm}

\hspace{1cm}
\begin{picture}(0,0)
    \put(-34,135){\rotatebox{0}{RGB}}
    \put(-34,95){\rotatebox{0}{GT}}
    \put(-34,55){\rotatebox{0}{Zoed}}
    \put(-34,15){\rotatebox{0}{Ours}}
\end{picture}
\includegraphics[width=0.93\textwidth]{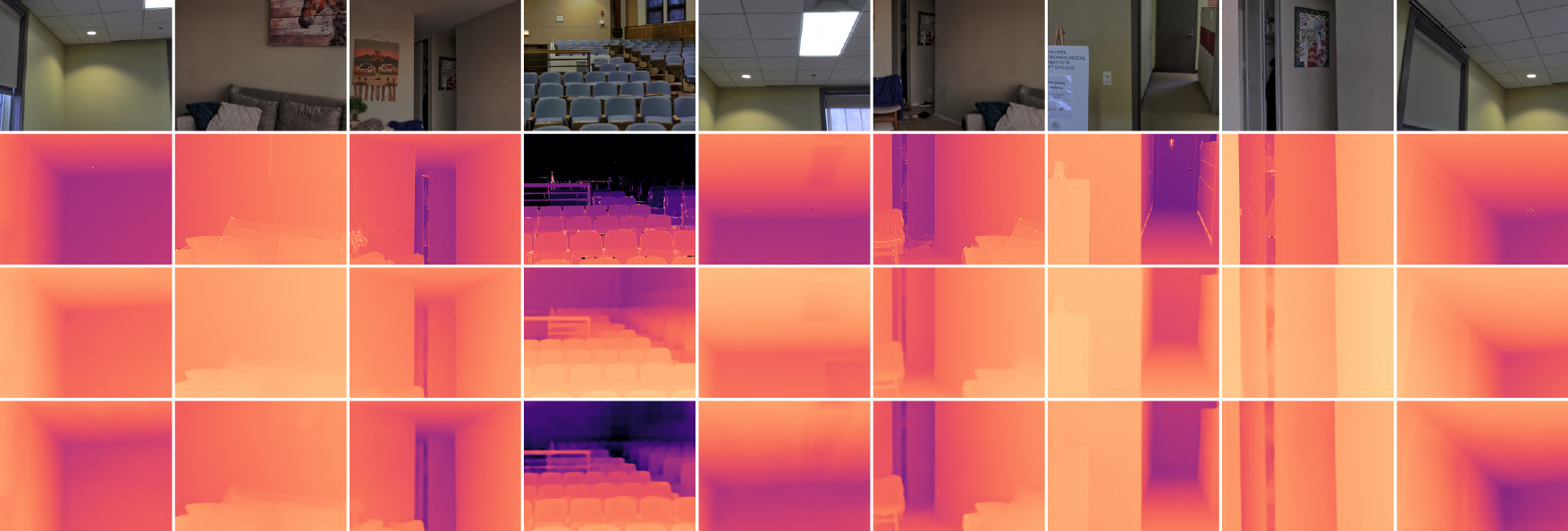}
\caption{\textbf{Qualitative Comparison on the \diode~\cite{diode_dataset} Dataset.} 
Our depth predictions are contrasted with those of Zoedepth\cite{zoedepth}. 
The first row displays RGB images, the second row shows groundtruth depth, 
the third row exhibits Zoedepth\cite{zoedepth}'s depth, and the fourth row showcases our depth predictions. 
To facilitate visual comparison, the colormap scale remains consistent across corresponding depth maps. 
Our model, trained only on \nyu, is compared with Zoedepth\cite{zoedepth}, which is trained on 12 datasets and then fine-tuned on \nyu.}
\label{fig:qualtitative_diode}
\end{figure*}

\begin{figure*}[h]
  \centering
  \hspace{1cm}
  \begin{picture}(0,0)
      \put(-34,135){\rotatebox{0}{RGB}}
      \put(-34,95){\rotatebox{0}{GT}}
      \put(-34,55){\rotatebox{0}{Zoed}}
      \put(-34,15){\rotatebox{0}{Ours}}
  \end{picture}
  \includegraphics[width=0.93\textwidth]{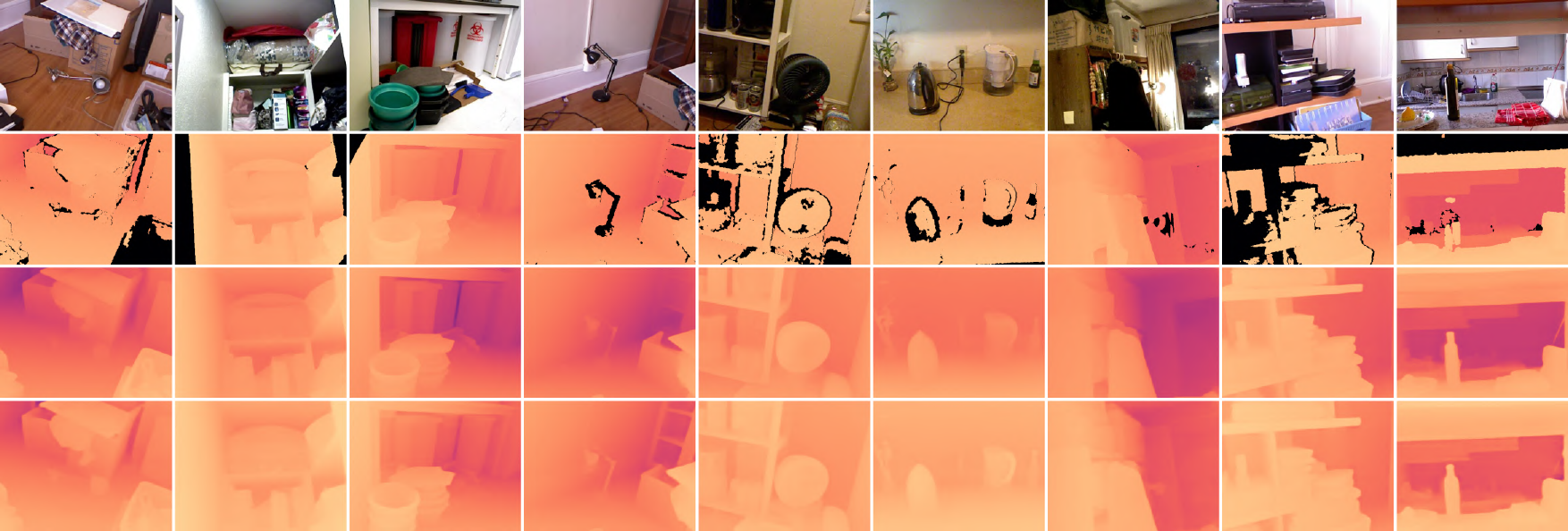}\vspace{1cm}
  
  \hspace{1cm}
  \begin{picture}(0,0)
      \put(-34,135){\rotatebox{0}{RGB}}
      \put(-34,95){\rotatebox{0}{GT}}
      \put(-34,55){\rotatebox{0}{Zoed}}
      \put(-34,15){\rotatebox{0}{Ours}}
  \end{picture}
  \includegraphics[width=0.93\textwidth]{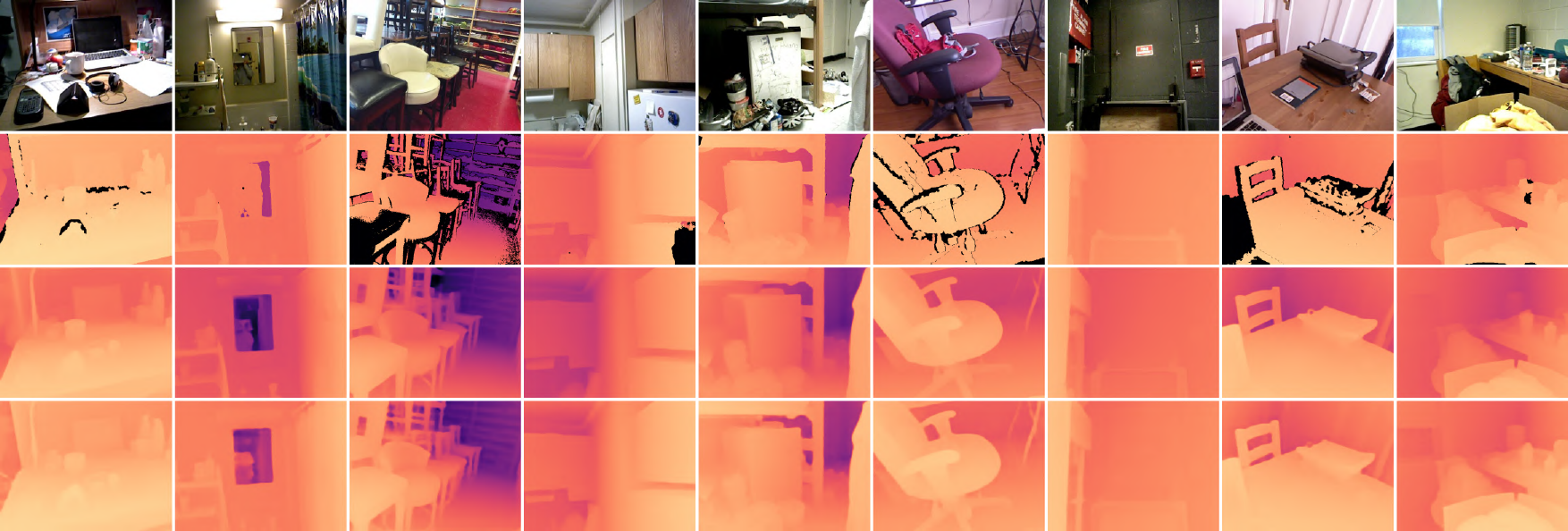}
  \caption{\textbf{Qualitative Comparison on the \sun~\cite{sunrgbd_dataset} Dataset.} 
Our depth predictions are contrasted with those of Zoedepth\cite{zoedepth}. 
The first row displays RGB images, the second row shows groundtruth depth, 
the third row exhibits Zoedepth\cite{zoedepth}'s depth, and the fourth row showcases our depth predictions. 
To facilitate visual comparison, the colormap scale remains consistent across corresponding depth maps. 
Our model, trained only on \nyu, is compared with Zoedepth\cite{zoedepth}, which is trained on 12 datasets and then fine-tuned on \nyu.}
  \label{fig:qualtitative_sun}
  \end{figure*}

\begin{figure*}[h]
  \centering
  \hspace{1cm}
  \begin{picture}(0,0)
      \put(-34,135){\rotatebox{0}{RGB}}
      \put(-34,95){\rotatebox{0}{GT}}
      \put(-34,55){\rotatebox{0}{Zoed}}
      \put(-34,15){\rotatebox{0}{Ours}}
  \end{picture}
  \includegraphics[width=0.93\textwidth]{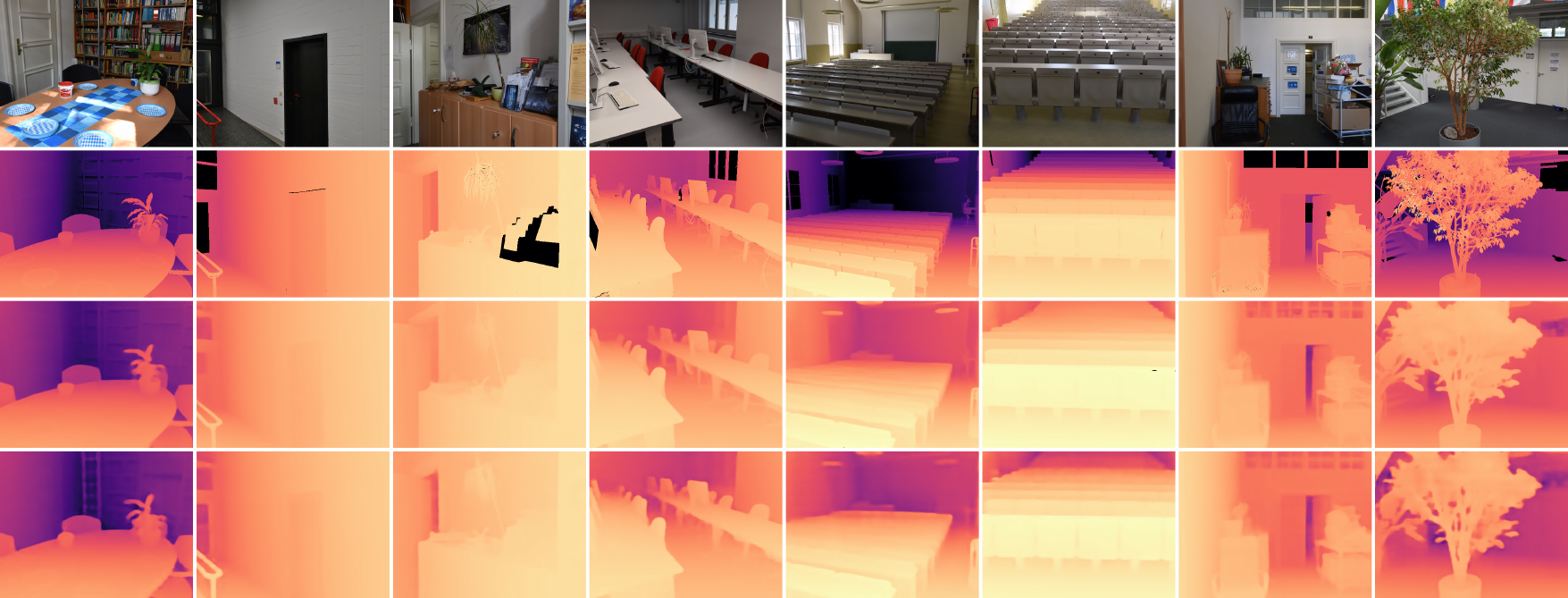}\vspace{1cm}
  
  \hspace{1cm}
  \begin{picture}(0,0)
      \put(-34,135){\rotatebox{0}{RGB}}
      \put(-34,95){\rotatebox{0}{GT}}
      \put(-34,55){\rotatebox{0}{Zoed}}
      \put(-34,15){\rotatebox{0}{Ours}}
  \end{picture}
  \includegraphics[width=0.93\textwidth]{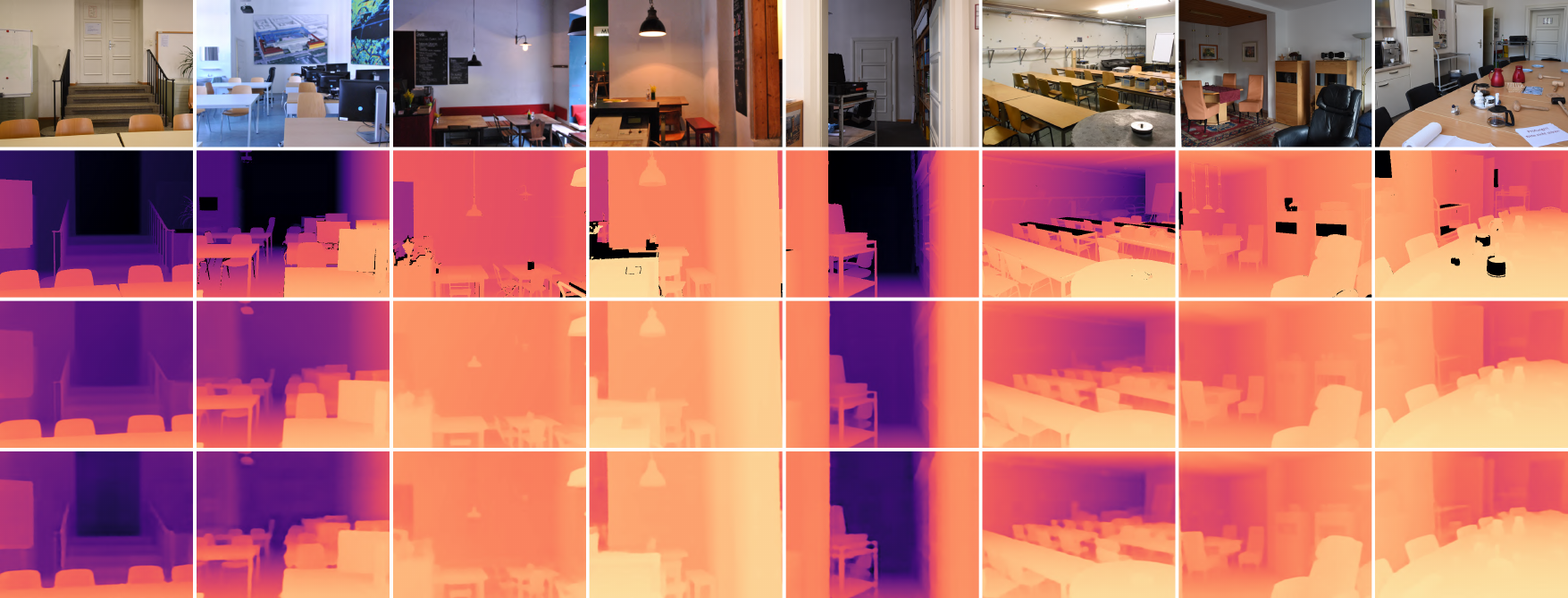}
  \caption{\textbf{Qualitative Comparison on the \ibims~\cite{ibims_dataset} Dataset.} 
Our depth predictions are contrasted with those of Zoedepth\cite{zoedepth}. 
The first row displays RGB images, the second row shows groundtruth depth, 
the third row exhibits Zoedepth\cite{zoedepth}'s depth, and the fourth row showcases our depth predictions. 
To facilitate visual comparison, the colormap scale remains consistent across corresponding depth maps. 
Our model, trained only on \nyu, is compared with Zoedepth\cite{zoedepth}, which is trained on 12 datasets and then fine-tuned on \nyu.}
  \label{fig:qualtitative_ibims}
  \end{figure*}